\documentclass[lettersize,journal]{IEEEtran}
\usepackage{amsmath,amsfonts}
\usepackage{algorithmic}
\usepackage{algorithm}
\usepackage{array}
\usepackage[caption=false,font=normalsize,labelfont=sf,textfont=sf]{subfig}
\usepackage{textcomp}
\usepackage{stfloats}
\usepackage{url}
\usepackage{verbatim}
\usepackage{graphicx}
\usepackage{cite}
\usepackage{booktabs}
\usepackage{makecell}
\usepackage[dvipsnames]{xcolor}
\usepackage{bm}
\usepackage{pifont}
\usepackage[table]{xcolor}
\usepackage{subcaption}
\usepackage{diagbox}
\usepackage{amssymb}
\usepackage{multirow}
\usepackage[breaklinks,colorlinks]{hyperref}
\usepackage{bbm}
\usepackage{pifont}
\newcommand{\cmarkg}{\ding{51}}
\newcommand{\xmark}{\ding{55}}
\newcommand{\xmarkg}{\textcolor{lightgray}{\ding{55}}}
\newcommand{\xmarkb}{\ding{55}}
\newcommand{\cmark}{\ding{51}}\newcommand{\ie}{\textit{i.e.}}
\newcommand{\eg}{\textit{e.g.}}

\definecolor{lightpink}{rgb}{1, 0.8, 0.9}
\begin{document}
\title{Referring Multi-Object Tracking with Comprehensive Dynamic Expressions}
\title{Referring Multi-Object Tracking with Diverse Dynamic Expressions}

\author{Yani Zhang$^{*}$, Dongming Wu$^{*}$, Wencheng Han, Xingping Dong, Shengcai Liao, Fellow, IEEE, and Bo Du
\IEEEcompsocitemizethanks{
\IEEEcompsocthanksitem Y. Zhang, X. Dong, and B. Du are with the School of Computer Science, National Engineering Research Center for Multimedia
Software, Institute of Artificial Intelligence, Hubei Key Laboratory of Multimedia and Network Communication Engineering, Wuhan University,
Wuhan 430072, China (Email: zebrazyn@whu.edu.cn, xingpingdong@whu.edu.cn, and dubo@whu.edu.cn).
D. Wu is with the Multimedia Lab (MMLab), Department of Information Engineering, the Chinese University of Hong Kong, Hong Kong, China (Email: wudongming97@gmail.com).
W. Han is with the School of Control Science and Engineering, Shandong University, Jinan, China (Email:  wencheng256@gmail.com).
S. Liao is with the Computer Vision Lab (CVLab), College of Computing and Artificial Intelligence (CCAI), United Arab Emirates University (UAEU), Al Ain, UAE.
(Email: scliao@ieee.org)
\IEEEcompsocthanksitem Corresponding author: \textit{Xingping Dong} and \textit{Bo Du}. $^{*}$ Equal contribution. This work was supported by National Natural Science Foundation of China under Grants 62225113 and 62471342, the Fundamental Research Funds for the Central Universities (No.~2042026kf0044), 
the Innovative Research Group Project of Hubei Province under Grants 2024AFA017, and the New Cornerstone Science Foundation through the XPLORER PRIZE.

}
\thanks{}}
\markboth{}{}

\maketitle
\begin{abstract}
Referring understanding is a fundamental task that bridges natural language and visual content by localizing objects described in free-form expressions. 
However, existing works are constrained by limited language expressiveness, lacking the capacity to model object dynamics in spatial numbers and temporal states.
To address these limitations, we introduce a new and general referring understanding task, termed referring multi-object tracking (RMOT). Its core idea is to employ a language expression as a semantic cue to guide the prediction of multi-object tracking, comprehensively accounting for variations in object quantity and temporal semantics.
Along with RMOT, we introduce the first RMOT benchmark, Refer-KITTI. We further present Refer-KITTI-V2 and Refer-Animal, which feature scalable and diverse language expressions.
To efficiently generate high-quality annotations covering object dynamics with minimal manual effort, we propose a semi-automatic labeling pipeline.
In addition, we propose TempRMOT, an elegant end-to-end  Transformer-based framework for RMOT. 
At its core is a query-driven Temporal Enhancement Module that represents each object as a Transformer query, enabling long-term spatial-temporal interactions with current objects and past frames to efficiently refine these queries. 
TempRMOT achieves state-of-the-art performance across multiple RMOT benchmarks, demonstrating the effectiveness of our approach.
The source code and dataset have been made publicly available at \href{https://github.com/zyn213/TempRMOT}{https://github.com/zyn213/TempRMOT}.
\end{abstract}

\begin{IEEEkeywords}
Referring Understanding, Referring Multi-Object Tracking,  Cross-modal Temporal Understanding
\end{IEEEkeywords}

\section{Introduction}
As an emerging task in computer vision, referring understanding~\cite{hui2023language, gao2023room,liang2023local,yang2020relationship,talk2car,khoreva2018video,nagaraja2016modeling,yu2016modeling,xie2025phrase,ke2025graph} aims to bridge the semantic gap between visual content and linguistic descriptions by locating and interpreting objects based on free-form textual queries. To advance this field, many datasets and benchmarks~\cite{young2014image,kazemzadeh2014referitgame,yu2016modeling,li2017tracking,vasudevan2018object,talk2car,khoreva2018video,seo2020urvos} have been proposed. However, previous benchmarks suffer from two typical limitations. \textbf{First}, each referring expression is typically annotated with only a single target, whereas in real-world scenarios, one expression can refer to multiple objects sharing the same semantic meaning.
\textbf{Second}, these benchmarks fail to model the temporal consistency between the language expression and the evolving state of the target. In many cases, an expression describes a transient state (\eg ``the turning cars''), but the annotation continues to track the object even after the described action is complete. 
Overall, current datasets do not provide reliable evaluation for scenarios involving multiple referent objects and temporal variations. 

\begin{figure}
\centering
\includegraphics[width=\linewidth]{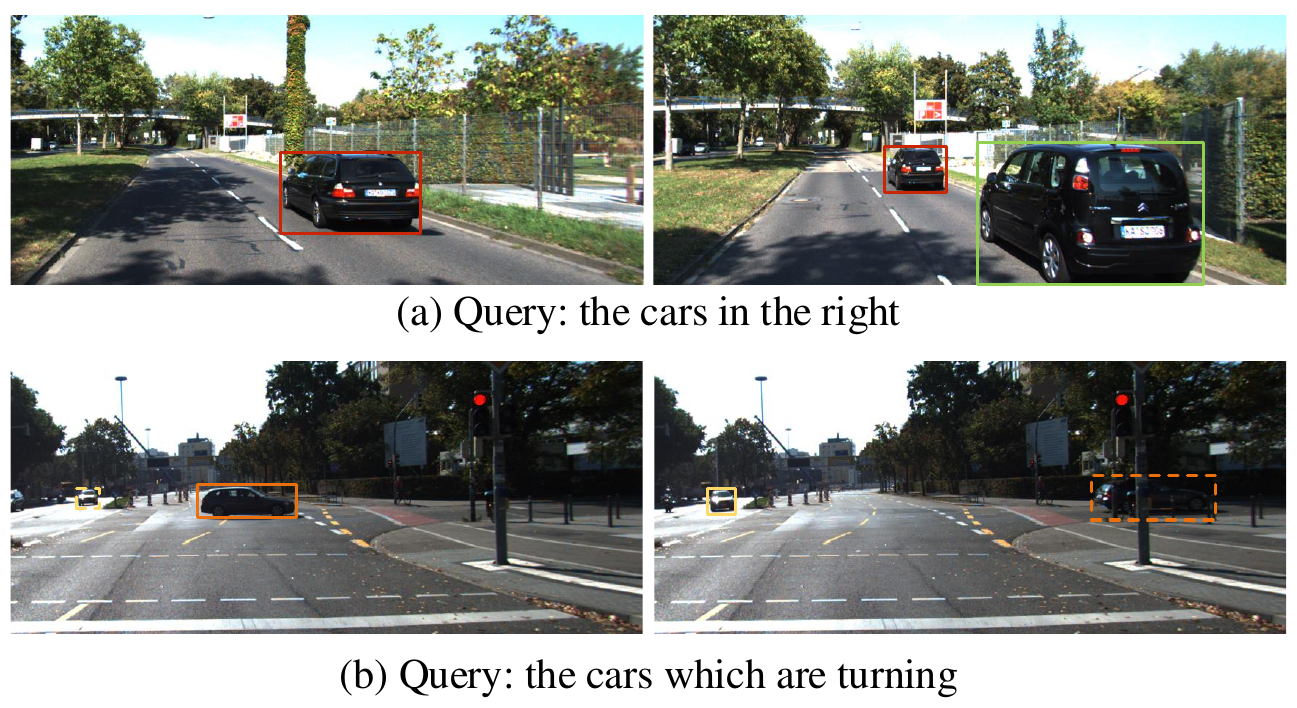}
\caption{\textbf{Representative examples from RMOT}. The expression query can refer to multiple objects of interest (a),  and captures the short-term status with accurate labels (b).}
\label{fig:motivation}
\vspace{-10pt}
\end{figure} 
\begin{figure*}
    \begin{minipage}[b]{.42\linewidth}
    \scriptsize
    \centering
    \small
    \begin{tabular}{l|cccc}
	\toprule
         \rowcolor[gray]{0.9}
        Dataset  &Video   &\makecell{Instances\\per Exp.}  &\makecell{Temporal.\\per Exp.}  \\
	\midrule
	RefCOCO~\cite{yu2016modeling}   & -   & 1    & 1    \\
	RefCOCO+~\cite{yu2016modeling}  & -   & 1    & 1    \\ 
	RefCOCOg~\cite{yu2016modeling}  & -   & 1    & 1    \\ 
	\midrule
	Talk2Car~\cite{talk2car} &\checkmark     &    1  & - \\ 
	VID-Sentence~\cite{chen2019weakly}     &\checkmark     & 1  & 1    \\ 
        Refer-DAVIS$_{17}$~\cite{khoreva2018video}&\checkmark & 1 & 1\\
        Refer-YV~\cite{seo2020urvos}&\checkmark & 1 &1\\
        \midrule
	Refer-KITTI                 			&\checkmark   &10.7   &  0.49\\ 
        \bottomrule
    \end{tabular} 
    \captionof{table}{\textbf{Comparison of Refer-KITTI with existing datasets.} Refer-YV means Refer-Youtube-VOS. `-' means unavailable. Exp. means Expressions}
    \label{tab:refer-kitt}
    \end{minipage}
    \  \   \   \   \   \
    \begin{minipage}[b]{.55\linewidth}
        \centering
        \includegraphics[width=9.5cm]{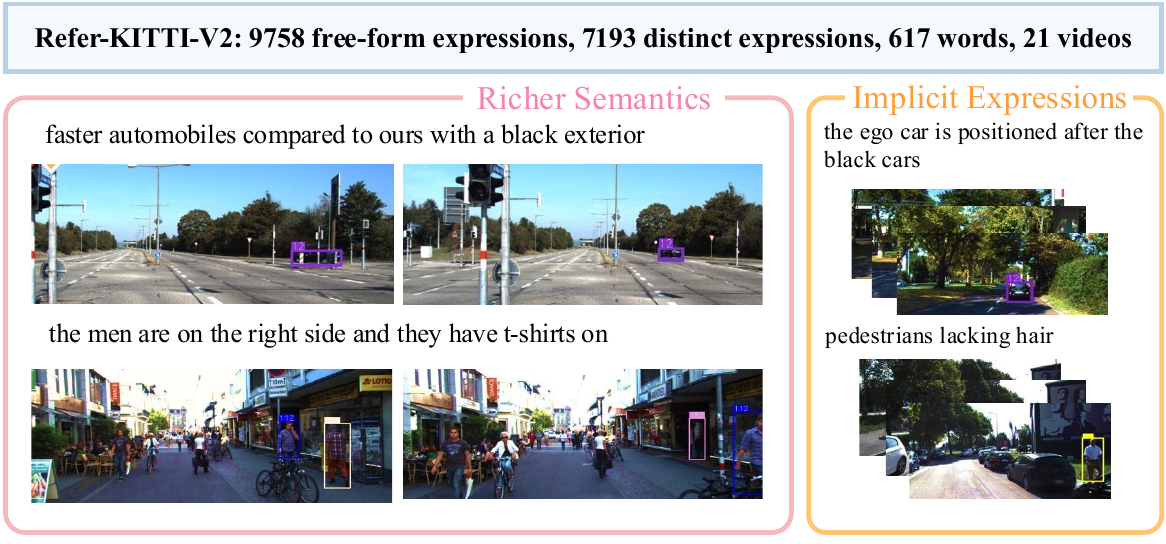}
        \caption{
            Refer-KITTI-V2 provides richer expressions
        from the perspective of numbers, semantics, and implicit descriptions.
        }
        \label{fig:refer-kitti-v2-rep}
    \end{minipage}
\vspace{-12pt}
\end{figure*}
To address these problems, we introduce a novel video understanding task guided by the language description, termed referring multi-object tracking (RMOT). Given a language expression, RMOT leverages it as a semantic cue to detect and track all described objects within a video.
Unlike previous tasks, RMOT better reflects real-world dynamic scenarios, as each expression can involve multiple variable objects.
For example, in Figure~\ref{fig:motivation}(a), the query ``the cars on the right'' corresponds to a variable number of targets, with one object in the $20^{th}$ frame and two objects in the $40^{th}$ frame. In addition, RMOT accounts for temporal state variations. As illustrated in Figure~\ref{fig:motivation}(b), for the query ``the cars which are turning'', objects are considered relevant only while performing the turning action and are disregarded once the turn is completed.

Along with RMOT, we also construct the first RMOT benchmark Refer-KITTI, built upon the public KITTI dataset~\cite{Geiger2012CVPR}. A thorough comparison between existing datasets and ours is summarized in Table~\ref{tab:refer-kitt}.
Compared to existing referring understanding datasets, Refer-KITTI offers two key advantages:
($i$) High flexibility in referent objects. The number of objects referenced per expression ranges from 0 to 105, with an average of 10.7, enabling diverse and complex tracking scenarios. 
($ii$) High temporal dynamics. Target objects exhibit long-duration motion across up to 400 frames. Our precise annotation tool effectively captures these temporal variations.

The RMOT task and Refer-KITTI have attracted significant research attention, inspiring the development of various models and datasets~\cite{wu2023referring,du2023ikun,lin2024echotrack,he2024visual,ma2024mls,wu2025language,liang2025cognitive,chen2025cross,zhao2025hff,liu2025few,chamiti2025refergpt,li2025nugrounding,li2025visual,davidson2025refav,huang2024tell,li2025cpany}. Despite this progress, existing datasets, such as Refer-KITTI~\cite{wu2023referring}, Refer-Dance~\cite{du2023ikun} and Refer-UE-City~\cite{ma2024mls}, are typically created through manual annotations combined with predefined language templates. This process not only requires significant human effort but also suffers from several limitations:
{1) Limited linguistic diversity. Language expressions are often constrained by fixed grammar templates, such as ``\textit{cars in \{color\}}''. These restrict linguistic diversity and realism, preventing the expressions from reflecting the richness of real-world language. 2) Semantic redundancy. As shown in Table~\ref{tab:datasets}, Refer-Dance~\cite{du2023ikun} contains 1,985 expressions but only 48 of them are unique, indicating a high level of repetition. Similar issues are observed in Refer-KITTI~\cite{wu2023referring}. 3) Lack of complex semantics. Existing expressions tend to focus on simple motions, such as ``walking'' or ``dancing''. They also rely heavily on direct visual attributes, for example, ``men with bags''. They lack the ability to represent negative descriptions or more complex semantics, such as ``men without bags''. These limitations motivate us to emphasize the importance of rich semantics for RMOT.}

To enrich language expressions with minimal effort, we propose a three-step semi-automatic labeling pipeline to efficiently generate annotations (see §III-A) and construct two larger and more diverse datasets, Refer-KITTI-V2 and Refer-Animal.
Our datasets have several appealing facets:
\ding{182} \textbf{Richer semantics}. 
In the first step of our three-step semi-automatic labeling pipeline, we annotate basic motion states such as ``walking" and ``moving", as well as the more complex motion states like ``speedier" and ``decelerating". This approach addresses the limitation of previous datasets that focused primarily on simple motions.
Besides, we integrate multiple basic aspects, e.g., category, color, and position, ensuring a holistic and detailed understanding of all visual objects.
Leveraging the strong language understanding and generation capabilities of large language models, our datasets highlight the richness and diversity of semantic information. For example, Refer-KITTI-V2 includes 7,193 distinct expressions and 617 unique words.
\ding{183} \textbf{Richer Number}. As shown in Table~\ref{tab:datasets}, Refer-KITTI-V2 contains a total of 9,758 expressions, which is 10× larger than Refer-KITTI. 
\ding{184} \textbf{Implicit Expressions}. Unlike previous RMOT datasets that focus solely on explicit descriptions to refer to objects, our dataset incorporates implicit instructions, such as ``the ego car is positioned after the black cars" and ``pedestrians lacking hair", to bridge the aforementioned gap. 

Although RMOT supports flexible referring scenarios, it poses challenges in multi-object prediction and temporal reasoning. Recent MOT methods, including Hybrid-SORT~\cite{yang2024hybrid}, SambaMOTR~\cite{segu2025samba}, and CO-MOT~\cite{yan2025co}, advance visual tracking through complementary association cues, synchronized state-space modeling, and improved query training, respectively. RMOT additionally requires interpreting object dynamics to ground motion-related expressions such as ``moving cars''. To address this challenge, we propose TempRMOT, an end-to-end RMOT framework that explicitly models long-term spatial-temporal dependencies. Built upon Deformable DETR~\cite{zhu2020deformable} and following the detect-and-track query formulation of MOTR~\cite{zeng2022motr}, TempRMOT introduces a Temporal Enhancement Module that retrieves historical object representations through cross-frame attention and subsequently integrates spatial interactions among current-frame objects. By jointly modeling object dynamics over time and inter-object spatial relations, this module enriches object queries with temporal evidence and spatial context for referring prediction and box refinement. Experiments on multiple RMOT benchmarks demonstrate the effectiveness of the proposed approach.

In summary, our main contributions are six-fold: 
\begin{itemize}
\item We introduce referring multi-object tracking (RMOT), a novel task that addresses the limitations of existing referring understanding by introducing dynamic multi-objects scenarios and temporal state variations.
\item We present the first RMOT benchmark, Refer-KITTI, to facilitate research in RMOT. To the best of our knowledge, Refer-KITTI is the first dataset specifically designed for predicting an arbitrary number of objects.
\item We further introduce Refer‑KITTI‑V2 and Refer‑Animal, which provide more diverse expressions across quantity, semantics, and implicit descriptions to emphasize the importance of rich semantics.
\item We propose an end-to-end baseline model for RMOT that supports tracking a variable number of referent objects while effectively modeling temporal state variations, without relying on any post-processing. 
\item We design a query-based Temporal Enhancement Module to explicitly model long-term spatial-temporal interactions, enabling progressive refinement of object representations over time.
\item TempRMOT achieves new state-of-the-art performance across multiple benchmarks, effectively handling dynamic scenes and complex referential expressions.
\end{itemize}

This work builds upon our conference paper~\cite{wu2023referring} and presents a substantial extension of it in various aspects. 
(1) We expand KITTI into a more diverse dataset, Refer-KITTI-V2, which overcomes the semantic limitations of template-based datasets. By introducing enriched keywords and free-form expressions, it better reflects real-world scenarios.
To broaden domain diversity, we additionally select 58 videos from AnimalTrack~\cite{zhang2023animaltrack} and construct Refer‑Animal following the same pipeline as Refer‑KITTI‑V2.
(2) We enhance the framework with a query-based temporal enhancement module (\S\ref{sec:tmn}) to explicitly model long-term spatial-temporal interactions, refining detection results using spatiotemporal information.
(3) We provide an in-depth discussion of our datasets (\S\ref{sec:statistics}), covering linguistic diversity, object distribution, the complexity of referring expressions, and a clearer taxonomy of expression types, offering valuable insights into the challenges inherent in RMOT.
(4) We propose a three-step semi-automatic labeling pipeline (\S\ref{sec:data_collection}) to annotate
the expressions with minimal human effort. It makes full
use of large language models to enhance both the quantity
and quality of the expressions generated.
(5) We conduct experiments on more benchmarks and compare our model with recent models~\cite{du2023ikun,lin2024echotrack,ma2024mls,he2024visual,zhuang2025cgatracker} (\S\ref{sec:sota}).
(6) We design more ablative studies (\S\ref{sec:ablation}) to quantitatively investigate core components of our model. We also provide additional visual results (\S\ref{sec:qulitative}) to show the  ability of TempRMOT with temporal dynamics and motion-related instructions
(7) We include a limitation analysis (\S\ref{sec:discussion}) and failure case visualizations (Figure~\ref{fig:failed_cases}) to highlight challenges and suggest future research directions.

\section{Related works}
\subsection{Referring Multi-Object Tracking Benchmark}

\begin{figure*}
  \centering
  \includegraphics[width=1\linewidth]{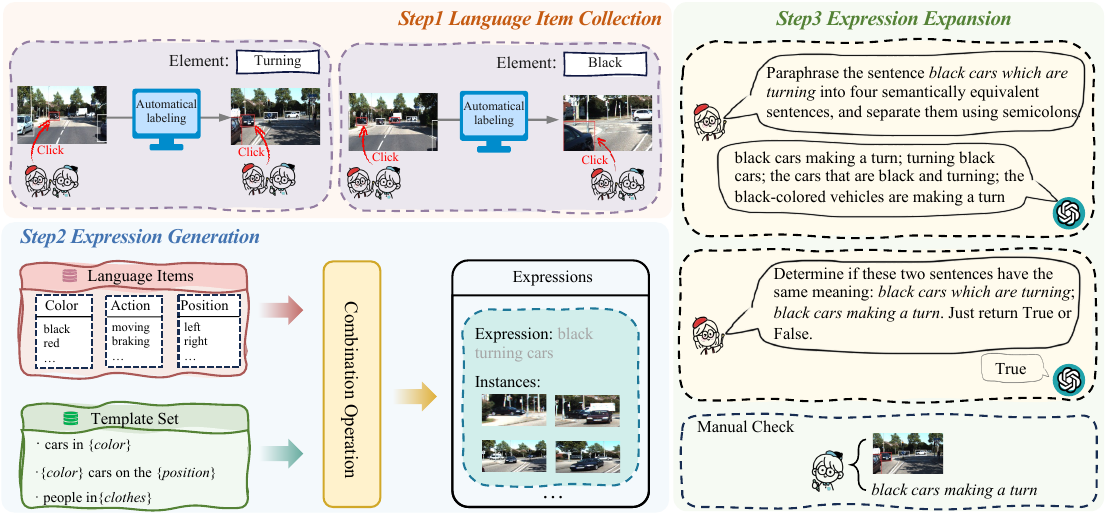}
  \caption{
  \textbf{Language Prompt Annotation Pipeline} consists of three steps: language item collection, prompt generation, and prompt expansion. Firstly, we use an efficient labeling tool to associate instances in each video with language elements at low human cost. Then, we manually create accurate language descriptions. 
  Finally, leveraging the powerful language understanding capabilities of large language models, we expand the new annotations with language descriptions.
  }
  \label{fig:dataset_pip}
\end{figure*}

Many advanced datasets have greatly contributed to the progress of referring understanding. Early efforts, including Flickr30k~\cite{young2014image}, ReferIt~\cite{kazemzadeh2014referitgame}, the ReferCOCO series~\cite{yu2016modeling}, Cops-Ref~\cite{chen2020cops}, focus on grounding concise natural language expressions to visual regions in static images. However, their static nature limits applicability to dynamic, real-world scenes. Recent video-based benchmarks~\cite{li2017tracking,vasudevan2018object,chen2019weakly,talk2car} address this by aligning language with visual content in videos. Referring understanding has also expanded to pixel-level video object segmentation~\cite{gavrilyuk2018actor,khoreva2018video,seo2020urvos,ding2023mevis}.
While these datasets have advanced the field, they are limited to single-object expressions and poorly capture temporal changes, restricting their effectiveness in dynamic multi-object scenarios.

To address these limitations, Wu \textit{et al.}~\cite{wu2023referring}  are the first to propose the task of referring multi-object tracking (RMOT), which extends referring understanding to dynamic, multi-object scenarios.
Unlike previous settings, RMOT accommodates expressions that may refer to multiple objects, a single object, or even none, thereby better reflecting the open-ended nature of real-world language grounding.
To support this new task, Wu \textit{et al.}\cite{wu2023referring} construct the first benchmark, Refer-KITTI, by augmenting the widely used multi-object tracking dataset KITTI\cite{Geiger2012CVPR} with referring expressions.
Building on this foundation, iKUN~\cite{du2023ikun} further proposes Refer-Dance, by enriching DanceTrack~\cite{sun2022dancetrack} with natural language descriptions.
Type-to-Track~\cite{nguyen2024type} introduces another large-scale RMOT benchmark GroOT based on multi-object tracking datasets that includes MOT17~\cite{milan2016mot16}, TAO~\cite{dave2020tao}, and MOT20~\cite{dendorfer2020mot20}.
Despite these promising developments, current RMOT benchmarks remain primarily constructed through manual annotation, rely on fixed rules for task formulation and emphasize simple movement rather than dynamic state changes.
As a result, they lack the diversity, flexibility, and scalability required to fully capture the richness and variability of language-grounded multi-object interactions.

\subsection{Referring Multi-Object Tracking Method}

Existing RMOT approaches can be broadly categorized into two families: two-stage methods~\cite{du2023ikun,li2025cpany,chamiti2025refergpt} and one-stage methods~\cite{wu2023referring,nguyen2024type,ma2024mls,li2025visual,zhuang2025cgatracker,li2025language,zhao2025hff}.
Two-stage methods generally involve two steps: they first extract object tracklets explicitly by leveraging existing multi-object tracking pipelines or specialized trackers, and then select those that match the given language expression. This design allows for more thorough refinement and filtering of potential candidates. However, their architectures are often complex and computationally expensive, which limits their practicality in real-world scenarios. In contrast, one-stage methods typically adopt an end-to-end Transformer-based framework. 
CGATracker~\cite{zhuang2025cgatracker} proposes a correlation-aware graph alignment method to better model the relationships between region-level and word-level features. 
ReferGPT~\cite{chamiti2025refergpt} introduces a zero-shot method for referring multi-object tracking. It uses a large language model and robust matching to track objects based on any open-set text query.
However, these methods primarily rely on adjacent-frame query propagation, resulting in insufficient modeling of long-term object dynamics. To address this limitation, TempRMOT introduces explicit long-term spatial-temporal modeling through historical query retrieval and inter-object refinement, providing direct access to temporal evidence and spatial context beyond propagated query representations.

\subsection{Query-based Temporal Modeling}
Employing query vectors to enhance temporal modeling has become a prevalent paradigm in Transformer-based architectures for both image ~\cite{li2024bevformer, zhou2022transvod, hou2023query, huang2021bevdet,li2023bevstereo} and video~\cite{cai2022memot,he2022inspro,wang2023exploring,qin2023motiontrack,li2023tcovis,hu2024temporal,gao2023memotr,yang2022temporally} understanding tasks. In multi-object tracking, MOTR~\cite{zeng2022motr} first introduces track queries to propagate object identities across frames, but it only considers adjacent frame attention and lacks long-term temporal aggregation. TransRMOT~\cite{wu2023referring} extends this idea to referring multi-object tracking with cross-modal fusion, but still can not handle long-term temporal aggregation.

Memory-based methods, such as MeMOT~\cite{cai2022memot} and MeMOTR~\cite{gao2023memotr}, leverage historical queries for temporal propagation.
ScoreMOT~\cite{zhao2024multi} improves association via motion prediction and hierarchical matching, while PeViTrack~\cite{li2025indoor} supports accurate, continuous indoor tracking through region search and memory.
However, these methods do not jointly model within-frame spatial interactions and language guidance.
In contrast, our approach leverages spatiotemporal information to explicitly refine detection through historical context and real-time spatial interactions, rather than simply injecting memory, making it both more effective and interpretable.
\section{Dataset Overview}
\label{sec:data_collection}
To push forward referring multi-object tracking, we first construct Refer-KITTI, which contains 895 referring expressions generated from predefined templates. 
To achieve this, we design an efficient labeling tool~\cite{wu2023referring}. Annotators select the target object by clicking its bounding box at the start and end of the action. The tool then automatically propagates labels to intermediate frames using the object ID. The resulting annotations, including frame ID, object ID, box coordinates, and the expression, are saved for training and evaluation.
Representative examples of Refer-KITTI are shown in Figure~\ref{fig:dataset_examples_rk}.
Although object IDs facilitate annotation, the scale and diversity of existing data remain limited. Therefore, we introduce two diverse benchmarks, Refer-KITTI-V2 and Refer-Animal, featuring multi-perspective natural language descriptions of object dynamics, and we develop a three-step semi-automatic pipeline for efficient, high-quality annotation with minimal manual effort (Figure~\ref{fig:dataset_pip}).

\subsection{Data Collection and Annotation}
\begin{figure*}
    \begin{minipage}[b]{.55\linewidth}
    \scriptsize
    \centering
    \small
    \begin{tabular}{l|ccccccc}
        \toprule
        Datasets&Videos&Word&Exp.&\makecell{Exp. per\\Video}&\makecell{Distinct\\Exp.}&\makecell{Implicit\\Exp.}\\
        \midrule
        
        Refer-KITTI~\cite{wu2023referring}&18&49&895&\makecell[c]{49.7}&215&\xmarkb\\
        GroOT$^*$~\cite{nguyen2024type}&14&260&1547&\makecell[c]{110.5}&1161&\xmarkb\\
        Refer-Dance~\cite{du2023ikun}&\textbf{65}&25&1985&\makecell[c]{30.5}&48&\xmarkb\\
        
        \midrule
        \textbf{Refer-Animal}&58&439&7810&\makecell[c]{146.8}&5712&\textbf{\cmark}\\
        \textbf{Refer-KITTI-V2}&21&\textbf{617}&\textbf{9758}&\makecell[c]{195.8}&\textbf{7193}&\textbf{\cmark}\\
        \bottomrule
        \end{tabular}
        \captionof{table}{Comparison of RMOT datasets. GroOT$^*$ represents the MOT17 subset with tracklet captions. ``Exp.'' means Expressions. ``Distinct exp.'' refers to the unique expression.}
        \label{tab:datasets}
    \end{minipage}
    \  \   \   \   \   \
    \begin{minipage}[b]{.4\linewidth}
        \centering
        \includegraphics[width=6.5cm]{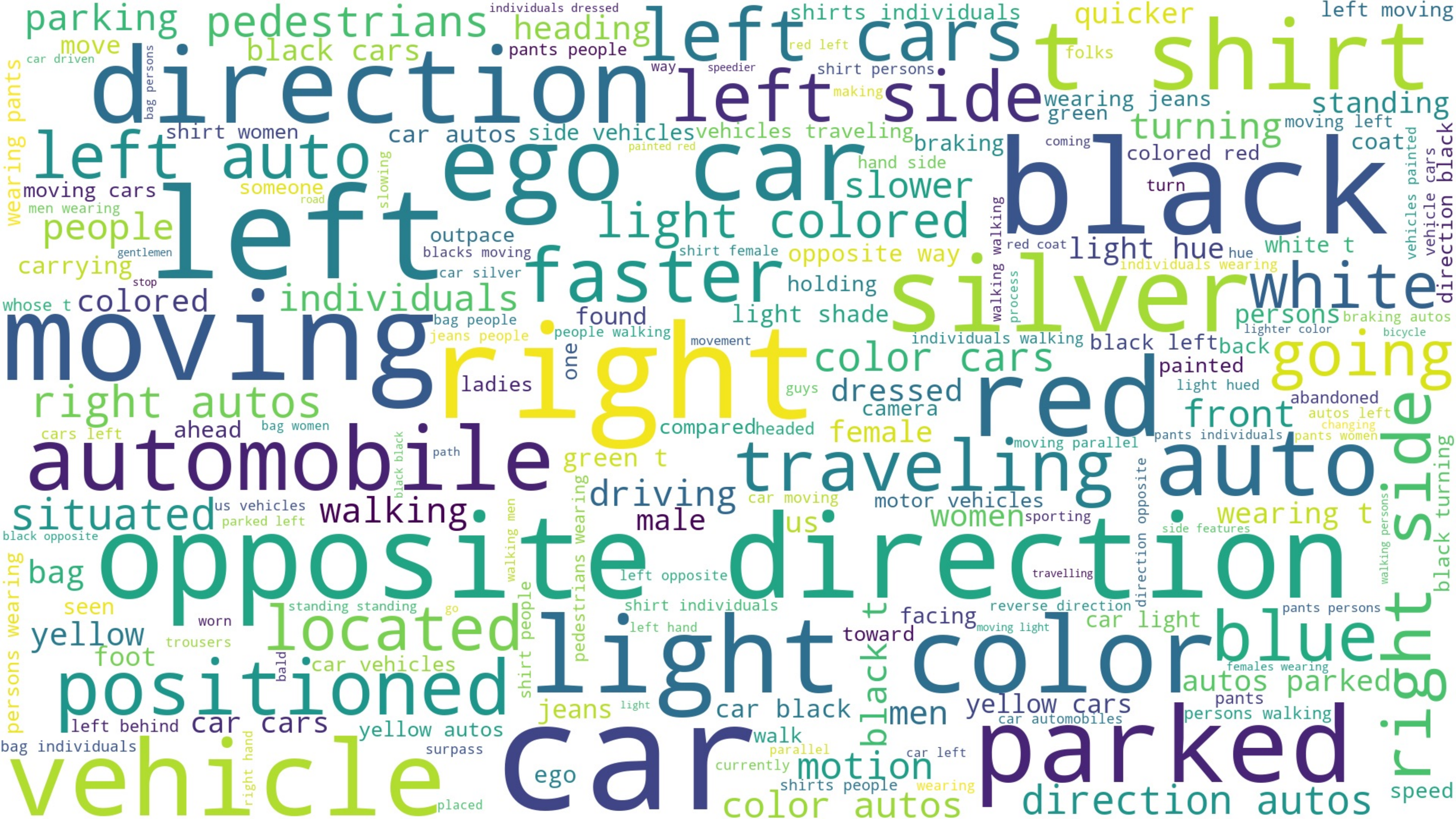}
        \caption{Word Cloud of all natural language expressions on our proposed Refer-KITTI-V2.}
        \label{fig:dataset-distinct}
    \end{minipage}
\vspace{-10pt}
\end{figure*}
\noindent \textbf{Step 1: Language Item Collection.}
We define a set of fundamental semantic attributes, called language items, to describe objects, including category (\textit{e.g.}, car, person), color (\textit{e.g.}, white, blue), and position (\textit{e.g.}, left, in front). For each video, we annotate these items as intermediate representations linking visual content and language. Using instance-level IDs from KITTI, annotators mark the start and end frames for each item, and labels are automatically propagated across frames. All annotations are manually verified for quality.
This results in structured annotations containing the frame ID, object ID, bounding box coordinates, and the associated language item.

\noindent \textbf{Step 2: Expression Generation.}
After obtaining the language items for each video, we apply predefined templates to generate initial expressions. These templates are designed to capture a wide range of semantic combinations, including formats such as ``\{\textit{class}\}-in-\{\textit{color}, \textit{position}, \textit{direction}\}" and ``\{\textit{class}\}-which-are-\{\textit{action}\}". For example, as shown in Figure~\ref{fig:dataset_pip},
we select the language items `turning' and `black', and match them with the template ``\{\textit{color}\}-\{\textit{action}\}-cars'' to produce the expression ``black turning cars''. To ground the expression in the visual domain, we perform an AND operation to associate the selected attributes with the intersecting bounding boxes in the video frames, thus forming a precise object set referred to by the expression.
To maintain the semantic integrity and visual grounding quality of the dataset, we manually curate attribute combinations by filtering out those with insufficient corresponding bounding boxes or ambiguous visual references. 
This process ensures expression accuracy as well as provides a reliable foundation for subsequent expression expansion.

\noindent \textbf{Step 3: Expression Expansion.}  
After Step 2, we collect a total of 2,719 initial language expressions. To further enrich linguistic diversity, we leverage a powerful large language model (LLM) to generate a wider variety of paraphrased expressions while maintaining the original meaning. Specifically, for each expression from Step 2, we use GPT-3.5~\cite{gpt} to generate four alternative variations. This process significantly expands the dataset, introducing varied linguistic patterns. The prompt guiding this generation process is shown in Figure~\ref{fig:dataset_pip}.

\textbf{Quality control for generated expressions.}
To ensure semantic fidelity and contextual relevance, we adopt a rigorous two-stage validation process. First, GPT-3.5 is tasked with verifying that each generated expression maintains the original meaning. Second, three independent human annotators manually review the outputs, filtering out expressions that are ambiguous or inconsistent with the visual content of the corresponding video clips. This human–LLM consistency check guarantees that only semantically faithful and visually grounded expressions are retained. Through this augmentation process, we substantially enrich the expression pool, expanding the total number of referring expressions to 9,758. The two-stage validation ensures that all expanded expressions are reliable, thereby providing a more diverse yet trustworthy linguistic resource for training and evaluation.

\subsection{Dataset Statistics}
\label{sec:statistics}
In Table~\ref{tab:datasets}, we compare Refer-KITTI-V2 and Refer-Animal with the existing RMOT dataset. 
The comparison demonstrates that Refer-KITTI-V2 contains the largest number of natural language expressions, as well as the highest diversity in both distinct expressions and unique word usage, reflecting its rich semantic variability.
Next, we discuss how  Refer-KITTI-V2 is deliberately constructed to increase the complexity of language-conditioned video understanding by introducing challenges across both linguistic and visual modalities.

\noindent \textbf{Video Content.} 
Our Refer-KITTI-V2 is built on one of the most popular datasets for multi-object tracking, KITTI~\cite{Geiger2012CVPR}, which provides 21 high-resolution and long temporal videos for training. Unlike Refer-KITTI, which abandons three over-complex videos and only uses the remaining 18 videos to annotate, we use all 21 videos to formulate more challenging datasets, with the maximum count reaching up to 1,059 frames. These deliberate design choices enhance the complexity and difficulty of Refer-KITTI-V2, posing greater challenges for language-conditioned video understanding tasks. 

\begin{figure*}[t]
\begin{center}
    \includegraphics[width=1 \linewidth]{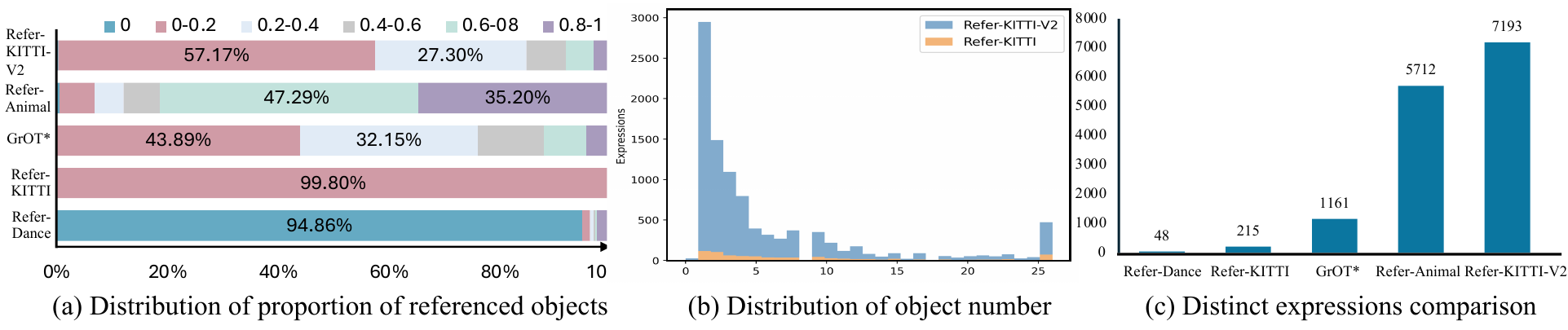} 
\end{center}
\caption{Statistics of Refer-KITTI-V2 and Refer-Animal. (a) Distribution of the proportion of referenced objects.
For example, in Refer-KITTI-V2, 57.17\% of the expressions refer to objects that make up 0-20\% of the total objects in the video.
(b) Distribution of object number. (c) Distinct expressions comparison.}
\label{fig:referent_objects}
\vspace{-5mm}
\end{figure*}

\noindent \textbf{Referent Objects.} 
Given that different datasets feature distinct visual scenes, we use the proportion of referenced objects to investigate the ratio of the objects referenced by an expression to the total number of objects in the scene. This metric helps determine whether the expression effectively utilizes the objects within the scene.
As shown in Figure~\ref{fig:referent_objects}(a), we visualize the distribution of this proportion across different datasets.
Our dataset and Refer-KITTI stand out as the expressions cover a wide range of object counts. This contrasts with Refer-Dance~\cite{du2023ikun}, where 94.86\% of expressions are null, and GroOT~\cite{nguyen2024type}, where expressions on average refer to one object. Additionally, we have included expressions unrelated to the video content to increase the benchmark difficulty, a feature absent in Refer-KITTI. This highlights the superior coverage and complexity of our expressions.  

Figure~\ref{fig:referent_objects}(b) further demonstrates that Refer-KITTI-V2 covers a significantly wider range of object instances compared to Refer-KITTI. Expressions in our dataset reference between 1 and 25 objects in most cases, with some reaching up to 105 instances. On average, each expression in the video contains approximately 6.69 objects. By fully utilizing the annotated objects, we have increased the complexity of the task.

\noindent \textbf{Language Expression.} 
As shown in Table~\ref{tab:datasets}, Refer-KITTI-V2 features the highest numbers of expressions, distinct expressions, and word counts among existing RMOT datasets. Additionally, it also encompasses several implicit expressions. We also provide the word cloud of Refer-KITTI-V2 in Figure~\ref{fig:dataset-distinct}. The word cloud illustrates that Refer-KITTI-V2 includes numerous words such as ``moving" and ``faster", which describe motion information, as well as ``black" and ``red", which describe the appearance and position of objects.

\noindent \textbf{Quality Assessment for LLM Paraphrase.}
We introduce a method for assessing LLM paraphrase quality, using the Refer-KITTI-V2 construction as an example. We first generate 10,041 distinct candidate expressions using GPT and then perform two-stage verification. Overall, 4.2\% contain semantic errors, 16.9\% are ambiguous, and 8.8\% are unnatural. All such expressions are discarded. Specifically, during manual filtering, three annotators independently review all expressions, achieving a Fleiss' Kappa~\cite{fleiss1971measuring} of 0.81 and confirming high consistency. The rejected expressions during manual filtering fall into three categories: ambiguous (22.1\%), unnatural (71.3\%), and incorrect (6.6\%).
For the retained LLM-generated expressions, the distribution of semantic attributes (\eg, color, action, position) remains proportional to human-written expressions, indicating no significant semantic bias.

\begin{figure}[t]
	\centering
	\includegraphics[width=1\linewidth]{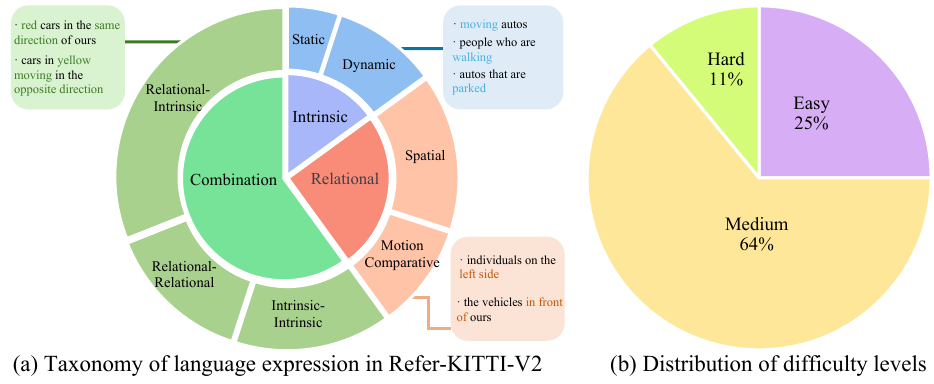}
	\caption{
    \textbf{The statistics of expressions in Refer-KITTI-V2.}
	}
	\label{fig:taxonomy}
\vspace{-8pt}
\end{figure}

\noindent \textbf{Taxonomy of Expression.} 
As shown in Figure~\ref{fig:taxonomy}, we categorize expressions based on their dominant grounding cue into three primary types, namely Relational, Intrinsic, and Combination. These are defined as grounded via relationships (\eg, with ego-vehicle), inherent properties (\eg, color or action), or a mix of both, respectively. Each type is further subdivided. Intrinsic covers Static (\eg, color) and Dynamic (\eg, moving) cues. Relational covers Spatial (position relative to ego-vehicle) and Comparative (\eg, speed comparison) cues. And Combination encompasses multiple intrinsic, multiple relational, or mixed cues. Based on this taxonomy, we define three difficulty levels: Easy, Medium, and Hard. Easy includes all Intrinsic and Relational expressions, Medium includes Combination expressions with fewer than three cues, and Hard includes Combination expressions with at least three cues, as well as Implicit or Negative expressions. The distribution across these levels is 25\% Easy, 64\% Medium, and 11\% Hard.
\section{Method}
\label{sec:method}
The overall architecture of our method is illustrated in Figure~\ref{fig:framework}. It builds on Transformer-based RMOT model~\cite{wu2023referring} and is further improved by a simple yet effective temporal enhancement module.
\subsection{Transformer-Based RMOT Model}
\label{sec:rmot}
The model formulates object representations as learnable queries, which are iteratively refined through a decoder by attending to cross-modal features.
Following MOTR~\cite{zeng2022motr}, the queries can be categorized into two types: those used for representing new targets in the current frame and those representing instance objects from the previous frame.

\noindent\textbf{Feature Extractor.}
Given a video sequence consisting of $T$ frames, we employ a convolutional neural network (CNN) backbone to extract multi-scale spatial features from each frame.
Specifically, for the $t$-th frame, the $l$-th level of the resulting feature pyramid is denoted as $\bm{I}_t^l \in \mathbb{R}^{C_l \times H_l \times W_l}$, where $C_l$, $H_l$, and $W_l$ correspond to the number of channels, height, and width of the feature map at scale $l$, respectively.
Meanwhile, a pre-trained language model encodes the input text with $L$ words into a sequence of word embeddings $\bm{S} \in \mathbb{R}^{L \times D}$, where $D$ is the embedding dimension.

\noindent\textbf{Cross-modal Encoder.}
\begin{figure*}[t]
  \centering
  \includegraphics[width=1\linewidth]{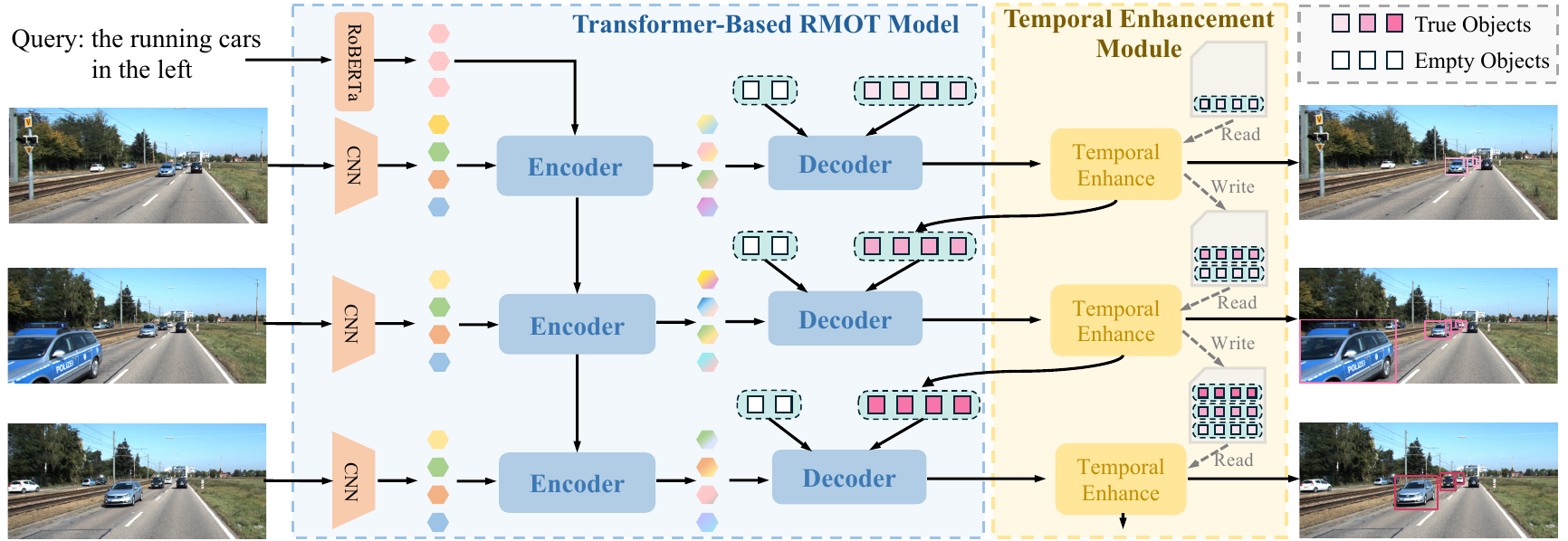}
  \caption{\textbf{Overall architecture of our TempRMOT.} The proposed framework comprises two key components: a Transformer-based RMOT module and a temporal enhancement module. The Transformer-based RMOT accepts a video frame, a language expression, and a set of learnable queries as input. Its output comprises a set of embeddings corresponding to the objects. The temporal enhancement module harnesses a memory mechanism to integrate temporal information for query enhancement. Our loss includes detection loss $\mathcal{L^I}$, based on image-level detection and spatio-temporal loss $\mathcal{L^S}$ based on temporal modeling.}
  \label{fig:framework}
\end{figure*}
The cross-modal encoder is responsible for accepting the visual and linguistic features and fusing them. 
The common strategy is to concatenate two kinds of features and feed them into the encoder to model dense connections via self-attention, like MDETR~\cite{mdetr}. 
However, the computation cost of self-attention is enormous due to the large token number of images.
To address this problem, we propose an early-fusion module to integrate the visual and linguistic features before deformable encoder layers.

Specifically, given the $l^{th}$ level feature maps $\bm{I}_t^l$, we use a $1\!\times\! 1$ convolution to reduce its channel number to $d\!=\!256$, and flatten it into a 2D tensor $\bm{I}_t^l\!\in\! \mathbb{R}^{H_lW_l \times d}$. 
To keep the same channels with visual features, the linguistic features are projected into $\bm{S}\!\in\! \mathbb{R}^{L \times d}$ using a fully-connected layer.
Three independent fully-connected layers transform the visual and linguistic features as $\bm{Q}$, $\bm{K}$, and $\bm{V}$:
\begin{equation}
	\small
	\vspace{-3pt}
	\begin{aligned}
	\bm{Q} &= \bm{W}_q (\bm{I}_t^l + P^V) \in \mathbb{R}^{H_lW_l \times d} , \\
	\bm{K} &= \bm{W}_k (\bm{S} + P^L)  \in \mathbb{R}^{L \times d},\\
	\bm{V} &= \bm{W}_v \bm{S}  \in \mathbb{R}^{L \times d},
	\end{aligned}
	\vspace{-1pt}
\end{equation}
where $\bm{W}$s are weights. $P^V$ and $P^L$  are position embedding of visual and linguistic features following~\cite{carion2020end,vaswani2017attention}.
We make matrix product on $\bm{K}$ and $ \bm{V}$,  and use the generated similarity matrix to weight linguistic features, \ie, $({\bm{Q}\bm{K}^\top}/{\sqrt{d}} )\bm{V} $. Here, $d$ is the feature dimension.
The original visual features are then added with the vision-conditioned linguistic features to produce the fused features $\hat{\bm{I}}_t^l$:
\begin{equation}
\small
	\vspace{-3pt}
	\hat{\bm{I}}_t^l = \frac{\bm{Q}\bm{K}^\top}{\sqrt{d}} \bm{V} + \bm{I}_t^l \in \mathbb{R}^{H_lW_l \times d}.
\end{equation}
After fusing two modalities, a stack of deformable encoder layers is used to promote cross-modal interaction:
\begin{equation}
\small
	\bm{E}_t^l = \text{DeformEnc}(\hat{\bm{I}}_t^l) \in \mathbb{R}^{H_lW_l \times d},
\end{equation}
where $\bm{E}_t^l$ is encoded cross-modal embedding, which will facilitate referring prediction in the following decoder.

\noindent\textbf{Decoder.}
The original decoder in the DETR framework uses learnable queries to probe encoded features for yielding instance embedding, further producing instance boxes and classes.
To associate objects between adjacent frames, we make full use of the decoder embedding from the last frame,  which is updated as  \textit{track query} of the current frame to track the same instance.
For new-born objects in the current frame, we adapt the original query from DETR, named \textit{detect query}. 
The tracking process is shown in Figure~\ref{fig:framework}.

Formally, let $\bm{D}_{t\!-\!1}\!\in\!\mathbb{R}^{N_{t\!-\!1}\!\times\! d}$ denote the  decoder embedding from the $(t\!-\!1)^{th}$ frame,  which is further transformed into \textit{track query} of  the $t^{th}$ frame, \ie, $Q^{tra}_{t}\!\in\!\mathbb{R}^{N_{t\!-\!1}'\!\times\! d}$, using self-attention and feed-forward network (FFN).  
Note that part of the $N_{t\!-\!1}$  decoder embeddings correspond to empty or exit objects, so we filter out them and only keep $N_{t\!-\!1}'$ true embeddings to generate  \textit{track query} $Q^{tra}_{t}$ in terms of their class score.
Let $Q^{det}\!\in\!\mathbb{R}^{N\!\times\! d}$ denote \textit{detect query}, which is randomly initialized for detecting new-born objects.
In practice, the two kinds of queries are concatenated together and fed into the decoder to learn target representation $\bm{D}_t$:
\begin{equation}
\small
	\bm{Q}_t = \text{Decoder}(\bm{E}_t^l, \text{concat}(Q^{det}, Q^{tra}_t)) \in \mathbb{R}^{N_t\!\times\! d},
\end{equation}
where the number of output embedding is $N_t\!=\!N_{t\!-\!1}' + N$, including track objects and detect objects.

\noindent\textbf{Referent Head.} 
After a set of decoder layers, we add a referent head on top of the decoder. The referent head includes class, box and referring branches.
The class branch is a linear projection, which outputs a binary probability that indicates whether the output embedding represents a true or empty object. The box branch consists of a three-layer feed-forward network with ReLU activations applied to the first two layers, which regresses the bounding box coordinates of visible instances.
Another linear projection acts as the referring branch to produce referent scores with binary values. It refers to the likelihood of whether the instance matches the expression.

\begin{figure}[t]
	\centering
	\includegraphics[width=1\linewidth]{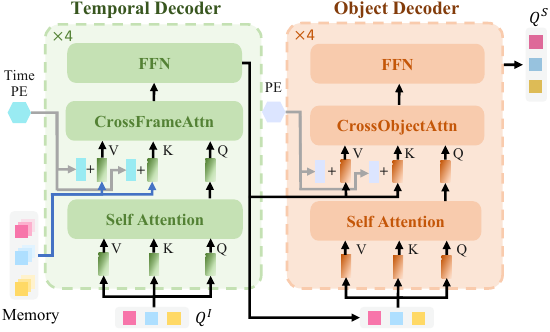}
	\caption{\textbf{The details of temporal enhancement module.} It contains two parts: temporal decoder and object decoder.
	}
	\label{fig:encoder}
\vspace{-8pt}
\end{figure}
\subsection{Temporal Enhancement Module}
\label{sec:tmn}
Existing methods typically propagate track queries between consecutive frames to implicitly preserve history. However, grounding motion-related expressions in RMOT requires observations across time, which may not be sufficiently represented by a single propagated query.
We therefore introduce a Temporal Enhancement Module that retains historical object queries, allowing the current query to directly access and selectively aggregate relevant information. Temporal positional encoding distinguishes observations across time to support reasoning about state changes, while interactions among current objects model spatial relationships. The enhanced spatial-temporal representations further guide residual box refinement. We detail the module below.

\noindent \textbf{Query Memory.}
As shown in the right part of Figure~\ref{fig:framework}, we propose an $N\!\times\! K$ query memory to store the query sets from the previous moments, where $N$ represents the number of stored frames, and K is the number of objects stored per frame. This memory operates on a first-in, first-out (FIFO) principle: as new frame information is appended, the oldest data in the queue is discarded. 
The $N \times K$ query memory serves as a dynamic temporal cache, continuously updating to retain recent query embeddings and facilitate temporal consistency.
The FIFO mechanism ensures that the system's memory consumption remains constant, even as new information is continuously integrated, making it highly suitable for environments with limited computational resources.

\noindent \textbf{Temporal Decoder.}
As illustrated in Figure~\ref{fig:encoder}, the proposed Temporal Decoder comprises 4 layers. Each layer consists of three main components: a self-attention module, a cross-attention (denoted as ``CrossFrameAttn''), and a feedforward neural network.
At each layer, the input query features $\mathbf{Q}_t$ at timestamp $t$ are first refined via self-attention to capture intra-frame dependencies.
To incorporate temporal context, we employ CrossFrameAttn. It enables each object-level query to attend over its corresponding historical representations within a temporal window of $\tau_h$ frames. The update is formulated as:
\begin{equation}
\label{eq:cross_frame_attn}
\begin{split}
     {\textbf{Q}}_{t} = \mathbf{CrossFrameAttn}(& \text{Q}=\textbf{Q}_{t}, \\ &\text{K}=\textbf{Q}_{t-\tau_h:t}, \text{V}=\textbf{Q}_{t-\tau_h:t}, \\ & \text{PE}=\mathbf{Pos}(t-\tau_h:t)),
\end{split}
\end{equation}
where $\textbf{Pos}(t-\tau_h:t)$ encodes the temporal positions of each frame, allowing the model to distinguish features across time. Empty historical entries are excluded from the attention computation.
This cross-frame attention facilitates explicit temporal reasoning by integrating object-specific information across time.
By jointly attending to both the current query and its historical context, the Temporal Decoder effectively models long-term temporal dependencies, enabling more accurate object state tracking and semantic alignment over time.

\noindent \textbf{Object Decoder.} The Object Decoder adopts a network architecture similar to that of the Temporal Decoder, with the main difference being in the positional embedding. 
Specifically, while the Temporal Decoder focuses on capturing temporal dependencies, the Object Decoder is designed to refine query from a spatial perspective.
To this end, it leverages cross-object attention to enable contextual interaction among the set of object-level queries. Given the query $\textbf{Q}_{t}$ output by the Temporal Decoder, the Object Decoder updates them as:
\begin{equation}
\label{eq:cross_frame_attn}
\begin{split}
     {\textbf{Q}}_{t} = \mathbf{CrossObjectAttn}(& \text{Q,K,V}=\textbf{Q}_{t}, 
     \\ & \text{PE}=\mathbf{Pos}(O^{1:N_t})),
\end{split}
\end{equation}
where $\text{Pos}(O^{1:N_t})$ represents the object-level positional embeddings, following~\cite{zhang2023motrv2}.
This mechanism enables feature exchange across $N_t$ objects, enhancing the discriminability of each object representation by incorporating spatial context.
The sequential Temporal Decoder and Object Decoder lead to $\bm{Q}^S_t$ ($S$ is short for ``spatio-temporal''):
\begin{equation}
\small
    \bm{Q}^S_t =  \textbf{ObjectDecoder}\left(\textbf{TemporalDecoder}(\bm{Q}_t)\right).
\end{equation}

\noindent \textbf{Track Refinement.}
To enhance the quality of object localization, we perform track refinement based on the query features enriched with temporal and spatial context. As mentioned above, the Transformer-based RMOT framework initially regresses the object bounding boxes $\bm{B}_t$ from each query via a box prediction branch. We then introduce an additional refinement stage to further improve the accuracy of these predictions.
Specifically, a multi-layer perceptron (MLP) is applied to the refined query features $\bm{Q}^S_t$ to predict residual adjustments to object properties, including center offsets $(\Delta x, \Delta y)$, size $(w, h)$, and confidence score $s$. These predicted residuals are then used to update the initial boxes, producing the final refined bounding boxes $\bm{B}_t$:
\begin{equation}
\small
\bm{B}_t = \bm{B}_t + \textbf{MLP}(\bm{Q}^S_t).
\end{equation}
\subsection{Loss Functions}
\label{sec:loss}
For training, we decouple the overall objective into two parts: the intra-frame loss $\mathcal{L}^{I}$ and the spatio-temporal loss $\mathcal{L}^{S}$.

\noindent \textbf{Intra-frame Loss.} The intra-frame loss consists of two parts: track loss and detect loss, corresponding to previously tracked objects and newly detected objects, respectively. We denote the set of predictions from the $t^\text{th}$ frame as $N_t$, which includes $N'_{t-1}$ tracking objects and $N$ detection objects. The tracking predictions are matched to their ground truths in a one-to-one manner, while the detection predictions are set predictions (i.e., $N$ is larger than the number of actual new objects). Therefore, we first calculate the track loss using tracking prediction set $ \{{\bm{c}}^{tra}_{t,i}, {\bm{b}}^{tra}_{t,i}, {\bm{r}}^{tra}_{t,i}\}_{i=1}^{N_{t-1}'}$ and the ground-truth set 
$ \{\hat{\bm{c}}^{tra}_{t,i}, \hat{\bm{b}}^{tra}_{t,i}, \hat{\bm{r}}^{tra}_{t,i}\}_{i=1}^{N_{t\!-\!1}'}$ directly. Here, ${\bm{c}}^{tra}_{t,i}\!\in\!\mathbb{R}^1$ is a probability scalar indicating whether this object is visible in the current frame.
${\bm{b}}^{tra}_{t,i}\!\in\!\mathbb{R}^4$ is a normalized vector that represents the center coordinates and relative height and width of the predicted box. 
${\bm{r}}^{tra}_{t,i}\!\in\!\mathbb{R}^1$ is a referring probability between the instance and the language description. The track loss $\mathcal{L}_t^{tra}$ is obtained via one-to-one computation:
\begin{equation}
\begin{aligned}
\small
\vspace{-10pt}
     \mathcal{L}_t^{tra} = \sum_{i=1}^{N_{t\!-\!1}'} \left[\lambda_{cls}  \mathcal{L}_{cls} ({\bm{c}}^{tra}_{t,i}, \hat{\bm{c}}^{tra}_{t,i} ) \right.
    +  \mathcal{L}_{box} ({\bm{b}}^{tra}_{t,i}, \hat{\bm{b}}^{tra}_{t,i})  \\
    +  \lambda_{ref}\mathcal{L}_{ref}({\bm{r}}^{tra}_{t,i}, \hat{\bm{r}}^{tra}_{t,i})\left.\right].
    \vspace{-3pt}
    \label{eq:track_loss}
\end{aligned}
\end{equation}
The box loss is a combination of L1 loss $\mathcal{L}_{L_1}$ and generalized IoU loss $\mathcal{L}_{giou}$, \ie, $\mathcal{L}_{box}\!=\!\lambda_{L_1}\mathcal{L}_{L_1} + \lambda_{giou}\mathcal{L}_{giou}$. $\mathcal{L}_{cls}$ and $\mathcal{L}_{ref}$ are the focal loss~\cite{lin2017focal}.
$\lambda_{L_1}$, $\lambda_{giou}$, $\lambda_{cls}$, and $\lambda_{ref}$ are the corresponding weight coefficients.

The detect loss $\mathcal{L}_t^{det}$ is calculated by performing bipartite matching between the set of detection predictions at time step $t$, denoted as  $\bm{y}^{det}_t\! =\!  \{{\bm{c}}^{det}_{t,i}, {\bm{b}}^{det}_{t,i}, {\bm{r}}^{det}_{t,i}\}_{i=1}^{N}$, and the corresponding ground-truth annotations of newly appeared objects, represented as $\hat{\bm{y}}^{det}_t$. To establish the optimal correspondence between predictions and ground truth, we identify the best permutation $\delta \in P_n $ that minimizes the overall matching cost:
\begin{equation}
	\hat{\delta} = \mathop{\arg \min} \limits_{\delta \in P_n} \mathcal{L}_{match}(\bm{y}^{det}_{t}, \hat{\bm{y}}^{det}_{t}),
\end{equation}
where $\mathcal{L}_{match} = \mathcal{L}_{box} + \lambda_{cls} \mathcal{L}_{cls}$. After obtaining the best permutation $\hat{\delta}$ with the lowest matching cost, we compute:
\begin{equation}
\begin{aligned}
\small
     \mathcal{L}_t^{det} =& \sum_{i=1}^{N} \left[ \lambda_{cls}  \mathcal{L}_{cls} +\mathbbm{1}  \mathcal{L}_{box}  \!+\! \mathbbm{1}  \lambda_{ref}\mathcal{L}_{ref}\right],
\end{aligned}
\end{equation}
where $\mathbbm{1}$  refers to $\mathbbm{1}_{\{\hat{\bm{c}}^{det}_{t,i} \neq \varnothing\}}$. 
Eventually, the final intra-frame loss $\mathcal{L}^{I}$ is the summation of track loss and detect loss:
\begin{equation}
\small
\mathcal{L}^{I} =  \sum_{t=1}^T( \mathcal{L}_t^{tra} + \mathcal{L}_t^{det}).
\end{equation}
As the first frame has no previous frames, its track query is set to empty. In other words, we only use the detect query to predict all new objects in the first frame.

\noindent \textbf{Spatio-temporal Loss.} After the transformer-based rmot model, the temporal enhancement module further processes the predicted queries by aggregating historical information.
We reuse the matching result $\hat{\delta}$ obtained from the detection stage and apply the same supervision to the outputs of the temporal enhancement module. Specifically, the spatio-temporal loss is computed with the same formulation as the detection loss:
\begin{equation}
\small
    \mathcal{L^S} = \lambda_{cls}^S\mathcal{L}_{cls}^S + \lambda_{box}^S\mathcal{L}_{box}^S +  \lambda_{ref}^S\mathcal{L}_{ref}^S, 
\end{equation}
where $\mathcal{L}_{cls}^S$ are focal loss for classification, $\mathcal{L}_{box}^S$ are L1 loss and the generalized IoU loss for box regression and $\mathcal{L}_{ref}^S$ are focal loss for referring. $\mathcal{L}_{cls}^S$, $\mathcal{L}_{box}^S$ and $\mathcal{L}_{ref}^S$ are generated by $\bm{Q}^S$. The corresponding weights are $\lambda_{cls}^S$, $\lambda_{box}^S$, and $\lambda_{ref}^S$.

\noindent \textbf{Final Loss.} The total training objective is the sum of intra-frame loss $\mathcal{L}^I$ and spatio-temporal loss $\mathcal{L}^S$:
\begin{equation}
\small
    \mathcal{L}^{final} = \mathcal{L}^I + \mathcal{L}^S.
\end{equation}

\begin{table*}[h]
\centering
\small
\caption{
Comparison with state-of-the-art methods on Refer-KITTI. ``E" means ``End to End". $^\dagger$ represents results after frame correction. The best results are in bold.
}
\resizebox{0.92\textwidth}{!}{
\setlength\tabcolsep{10pt}
\begin{tabular}{l|l|cccccccc}
\toprule
\rowcolor[gray]{.9}
Method & E & HOTA & DetA & AssA & DetRe &	DetPr &	AssRe &	AssPr & LocA \\

\midrule
FairMOT$^\dagger$~\cite{zhang2021fairmot} &\xmarkb&22.78&14.43&39.11&16.44&45.48&43.05&71.65 & 74.77\\
DeepSORT~\cite{wojke2017simple}  &\xmarkb&25.59&19.76&34.31&26.38&36.93&39.55&61.05&71.34\\
ByteTrack$^\dagger$~\cite{zhang2022bytetrack}&\xmarkb&24.95&15.50&43.11&18.25&43.48&48.64&70.72&73.90\\
CStrack~\cite{liang2022rethinking} &\xmarkb&27.91&20.65	&39.10	&33.76	&32.61	&43.12	&71.82	&79.51\\
TransTrack~\cite{sun2020transtrack} &\xmarkb&32.77&23.31&45.71&32.33&42.23&49.99&78.74&79.48\\
TrackFormer~\cite{meinhardt2022trackformer}&\xmarkb&33.26&25.44&45.87&35.21&42.10&50.26&78.92&79.63\\
iKUN~\cite{du2023ikun}&\xmarkb&48.84&35.74&\textbf{66.80}&51.97&52.25&72.95&87.09&- \\
{ReferGPT}~\cite{chamiti2025refergpt}&\xmarkb&46.36&36.58&59.00&51.40&52.16&\textbf{73.16}&69.31&83.26 \\

\midrule
EchoTrack~\cite{lin2024echotrack}&\cmarkg&39.47&31.19&51.56&42.65&48.86&56.68&81.21&79.93 \\
DeepRMOT~\cite{he2024visual}&\cmarkg&39.55&30.12&53.23&41.91&47.47&58.47&82.16&80.49 \\
TransRMOT$^\dagger$~\cite{wu2023referring}&\cmarkg&46.56&37.97&57.33&49.69&\textbf{60.10}&60.02&89.67&90.33 \\
CGATracker~\cite{zhuang2025cgatracker}&\cmarkg&48.81&38.66&61.77&51.67&59.12&65.87&\textbf{90.26}&- \\
MLS-Track~\cite{ma2024mls}&\cmarkg&49.05&40.03&60.25&\textbf{59.07}&54.18&65.12&88.12&- \\
\rowcolor{pink!40}
\textbf{TempRMOT(Ours)}&\cmarkg&\textbf{52.21}&\textbf{40.95}&66.75&55.65&59.25&71.82&87.76&\textbf{90.40} \\
\bottomrule
\end{tabular} 
}
\label{tab:sota_kitti}

\end{table*}
\begin{table*}[]
\centering
\small
\caption{
Comparison with state-of-the-art methods on Refer-KITTI-V2 and KITTI. ``E" means ``End to End". The best results are in bold.
}
\resizebox{0.92\linewidth}{!}{
\setlength\tabcolsep{10pt}
\begin{tabular}{p{26mm}|l|cccccccc}
\toprule
\rowcolor[gray]{.9}
Method & E & HOTA & DetA & AssA & DetRe &	DetPr &	AssRe &	AssPr & LocA \\
\midrule
\multicolumn{ 10 }{l}{\textbf{\textit{Refer-KITTI-V2 dataset}}} \\
FairMOT~\cite{zhang2021fairmot}&\xmark&22.53& 15.80& 32.82& 20.60& 37.03& 36.21& 71.94& 78.28\\
ByteTrack~\cite{zhang2022bytetrack}&\xmark&24.59& 16.78& 36.63& 22.60& 36.18& 41.00& 69.63& 78.00\\
iKUN~\cite{du2023ikun}&\xmark&10.32& 2.17& 49.77& 2.36& 19.75& 58.48& 68.64& 74.56\\
ReferGPT~\cite{chamiti2025refergpt}&\xmark&30.12&15.69&\textbf{59.02}&21.55&34.41&\textbf{74.59}&68.20&79.76 \\
TransRMOT~\cite{wu2023referring} &\cmarkg&31.00&19.40&49.68&36.41&28.97&54.59&82.29&89.82 \\
CGATracker~\cite{zhuang2025cgatracker} &\cmarkg&33.19&22.04&50.13&\textbf{38.00}&33.88&54.73&\textbf{84.90}&- \\
\rowcolor{pink!40}
\textbf{TempRMOT(Ours)}&\cmark&\textbf{35.04}&\textbf{22.97}&53.58&34.23&\textbf{40.41}&59.50&81.29&\textbf{90.07} \\
\midrule
\multicolumn{ 10 }{l}{\textbf{\textit{Refer-Animal dataset}}} \\
FairMOT~\cite{zhang2021fairmot}&\xmark&16.72& 9.04& 31.02& 11.71& 27.88& 33.96& 81.34& 87.92  \\
ByteTrack~\cite{zhang2022bytetrack}&\xmark&18.13& 10.05& 32.83&11.77& \textbf{39.93}& 35.08& 88.52& 87.88 \\
iKUN~\cite{du2023ikun}&\xmark&5.21& 0.88& 31.20& 8.55& 0.98& 33.47& 76.49& 80.41 \\
TransRMOT~\cite{wu2023referring} &\cmarkg&23.26& 13.83& 39.34& \textbf{23.54}& 24.78& 42.45& 81.63& 88.71 \\
\rowcolor{pink!40}
\textbf{TempRMOT(Ours)}&\cmark&\textbf{27.37}& \textbf{17.36}& \textbf{43.35}& 23.48&39.23&\textbf{45.95}& \textbf{88.83}& \textbf{89.16}  \\
\midrule
\multicolumn{ 10 }{l}{\textbf{\textit{KITTI dataset}}} \\
TransRMOT~\cite{wu2023referring}&\cmarkg&61.52& \textbf{57.16}& 66.51& \textbf{64.19}& 81.23& 69.80& \textbf{91.60}&90.88\\
\rowcolor{pink!40}
\textbf{TempRMOT(Ours)}&\cmarkg&\textbf{63.47}&56.09&\textbf{72.04}&61.56&\textbf{83.68}&\textbf{76.07}&89.67&\textbf{91.19}\\
\bottomrule
\end{tabular} 
}
\label{tab:sota_refer_kitti_v2}
\end{table*}
\section{Experiments}

\label{sec:expre}
\subsection{Datasets and Evaluation Metrics}

\noindent \textbf{Dataset.}
For a comprehensive evaluation, we conduct experiments on six video datasets: Refer-KITTI~\cite{wu2023referring}, Refer-KITTI-V2, Refer-Animal, Refer-Dance~\cite{du2023ikun}, Refer-BDD~\cite{lin2024echotrack}, and KITTI~\cite{Geiger2012CVPR}. Refer-KITTI stands as the inaugural publicly available RMOT dataset. We construct Refer-KITTI-V2 by extending KITTI~\cite{Geiger2012CVPR} and construct Refer‑Animal based on the AnimalTrack~\cite{zhang2023animaltrack}. Refer-Dance~\cite{du2023ikun} and Refer-BDD~\cite{lin2024echotrack} are built upon DanceTrack~\cite{sun2022dancetrack} and BDD100k~\cite{yu2020bdd100k}, respectively.

\noindent \textbf{Evaluation Metric.}
We adopt Higher Order Tracking Accuracy (HOTA)~\cite{luiten2020IJCV} as standard metrics to evaluate the two benchmark, Refer-KITTI and Refer-KITTI-V2. 
Its core idea is calculating the similarity between the predicted and ground-truth tracklet.
Unlike MOT using HOTA to evaluate all visible objects, when those non-referent yet visible objects are predicted, they are viewed as false positives in our evaluation.
As the HOTA  score is obtained by combining  Detection Accuracy (DetA) and Association Accuracy (AssA), \ie, $\text{HOTA}\!=\!\sqrt{\text{DetA}\cdot \text{AssA}}$, it performs a great balance between measuring frame-level detection and temporal association performance. 
Here, DetA defines the detection IoU score, and AssA is the association IoU score.
In this work, the overall HOTA is calculated by averaging across different sentence queries.

\subsection{Implementation Details}
\noindent \textbf{Model Details.}
We adopt visual backbone ResNet-50~\cite{he2016deep} and text encoder RoBERTa~\cite{liu2019roberta} in our TransRMOT. 
Similar to Deformable DETR~\cite{zhu2020deformable}, the last three stage features $\{ \bm{I}^3_t, \bm{I}^4_t, \bm{I}^5_t\}$ from the visual backbone are used for further cross-modal learning. Besides, the lowest resolution feature map $\bm{I}^6_t$ is added via a $3\!\times\!3$ convolution with spatial stride 2 on the $\bm{I}^5_t$.
Each of the multi-scale feature maps is independently performed the cross-modal fusion. After that, deformable attention in the encoder and decoder integrates the multi-scale features.
The architecture and number of the encoder and decoder layer follow the setting of \cite{zhu2020deformable}.
The temporal enhancement module uses a memory length of $N=4$ for Refer-KITTI and $N=5$ for Refer-KITTI-V2.

\noindent \textbf{Training.}
Following the setting of TransRMOT~\cite{wu2023referring}, 
the parameters of Transformer-Encoder and Transformer-Decoder are initialized from the official Deformable DETR~\cite{zhu2020deformable} with iterative bounding track refinement weights pre-trained on the COCO dataset~\cite{lin2014microsoft}. 
Random crop is used for data augmentation.
The shortest side ranges from 800 to 1536 for multi-scale learning.
During training, the text encoder's parameters remain frozen, while all other parameters are randomly initialized. Our optimization strategy employs Adam with a base learning rate of 1e-5, except for the visual backbone, which uses a learning rate of 1e-5. Beginning from the 40th epoch, the learning rate is reduced by a factor of 10. 
Moreover, object erasing and inserting are added to simulate object exit and entrance following~\cite{zeng2022motr}.
The loss coefficients are set as: $\lambda_{cls}^D\!=\!5$, $\lambda_{L_1}^D\!=\!2$,  $\lambda_{giou}^D\!=\!2$, $\lambda_{ref}^D\!=\!2$,
$\lambda_{cls}^R\!=\!5$, $\lambda_{L_1}^R\!=\!2$,  $\lambda_{giou}^R\!=\!2$, $\lambda_{ref}^R\!=\!2$.
The entire network is trained end-to-end with a batch size of 1 for 60 epochs on 4 RTX 4090 GPUs.
\begin{table*}[t!]
\caption{Ablation studies of our proposed TempRMOT on Refer-KITTI-V2.}
\centering

\subfloat[
Ablation study on TempRMOT.
\label{table:module_effective}
]{
\begin{minipage}{0.32\linewidth}
\centering
\begin{tabular}{cc|ccc}
\rowcolor[gray]{.9}
\toprule
Temp. & Refine & HOTA & DetA & AssA \\
\midrule
\makecell[c]{\xmarkg} & \makecell[c]{\xmarkg} & 31.00 & 19.40 & 49.68 \\
\makecell[c]{\cmarkg} & \makecell[c]{\xmarkg} & 34.46 & 22.73 & 52.37 \\
\rowcolor{pink!40}
\cmarkg & \cmarkg & \textbf{35.04} & \textbf{22.97} & \textbf{53.58} \\
\bottomrule
\end{tabular}
\end{minipage}
}
\hfil
\subfloat[
Different lengths for training.
\label{table:length_a}
]{
\begin{minipage}{0.32\linewidth}
\centering
\begin{tabular}{c|ccc}
\toprule
\rowcolor[gray]{.9}
$N_t$ & HOTA & DetA & AssA \\
\midrule
3 & 33.64 & 21.96 & 51.66 \\
4 & 34.41 & 22.43 & 52.90 \\
\rowcolor{pink!40}
\textbf{5} & \textbf{34.72} & \textbf{22.59} & \textbf{53.49} \\
\bottomrule
\end{tabular}
\end{minipage}
}
\hfil
\subfloat[
Different lengths for inference.
\label{table:length_b}
]{
\begin{minipage}{0.32\linewidth}
\centering
\begin{tabular}{c|ccc}
\toprule
\rowcolor[gray]{.9}
$N_i$ & HOTA & DetA & AssA \\
\midrule
5 & 34.72 & 22.59 & 53.49 \\
6 & 34.78 & 22.73 & 53.32 \\
\rowcolor{pink!40}
\textbf{8} & \textbf{35.04} & \textbf{22.97} & \textbf{53.58} \\
\bottomrule
\end{tabular}
\end{minipage}
}

\vspace{-3mm}
\end{table*}

\noindent \textbf{Inference.}
During the inference time, our model operates without the need of post-processing. 
At the  $t^{th}$ frame, it produces $N_t$ instance embeddings,  each corresponding to true or empty objects.
We choose these embeddings whose class score exceeds 0.6 to yield true object boxes, and referent objects are selected using a referring threshold $\beta_{ref} = 0.4$.

\subsection{State-of-the-art Comparison}
\label{sec:sota}
\noindent \textbf{Refer-KITTI.}
As shown in Table~\ref{tab:sota_kitti}, we compare our method with existing approaches on Refer-KITTI. Notably, we identified a misalignment between predicted and ground-truth frames in the original benchmark. To ensure fair comparison, we corrected the results from~\cite{wu2023referring}, including FairMOT~\cite{zhang2021fairmot}, ByteTrack~\cite{zhang2022bytetrack}, and TransRMOT~\cite{wu2023referring}. The corrected results are denoted as FairMOT$^\dagger$, ByteTrack$^\dagger$ and TransRMOT$^\dagger$, respectively. 
Compared with two-stage methods, which typically decouple detection and tracking, TempRMOT demonstrates a clear performance advantage, suggesting that jointly modeling spatiotemporal and linguistic cues is more effective in this task. In addition, while one-stage methods generally perform better in this benchmark due to their end-to-end optimization and tighter integration of components, TempRMOT still achieves the highest HOTA score among them. 
Specifically, it outperforms the previous best model, MLS-Track, by 3.16, setting a new state-of-the-art with a HOTA of 52.21. 

\noindent \textbf{Refer-KITTI-V2.}
In Table~\ref{tab:sota_refer_kitti_v2}, we report results on Refer-KITTI-V2. 
To make a fair comparison, we developed a series of CNN-based competitors by integrating our cross-modal fusion module into the detection component of multi-object tracking models,  like FairMOT~\cite{zhang2021fairmot} and ByteTrack~\cite{zhang2022bytetrack}. 
On this novel dataset, TempRMOT also achieves the state-of-the-art performance, surpassing the strong competitor TransRMOT~\cite{wu2023referring} by margins of 4.04\%, 3.57\%, and 3.9\% on HOTA, DetA, and AssA, respectively.
The results show the generalization and scalability of TempRMOT. 

Furthermore, the HOTA score achieved by TempRMOT on the Refer-KITTI-V2 dataset is lower than that on Refer-KITTI. This performance degradation is primarily attributed to the increased linguistic complexity introduced in Refer-KITTI-V2, where more diverse and semantically enriched expressions are adopted. While such variations enhance the descriptive power of the language, they also impose greater challenges to accurately associate textual cues with visual instances.
 \begin{figure*}[t]
	\centering
	\resizebox{\textwidth}{!}
	{
		\includegraphics[width = 16cm]{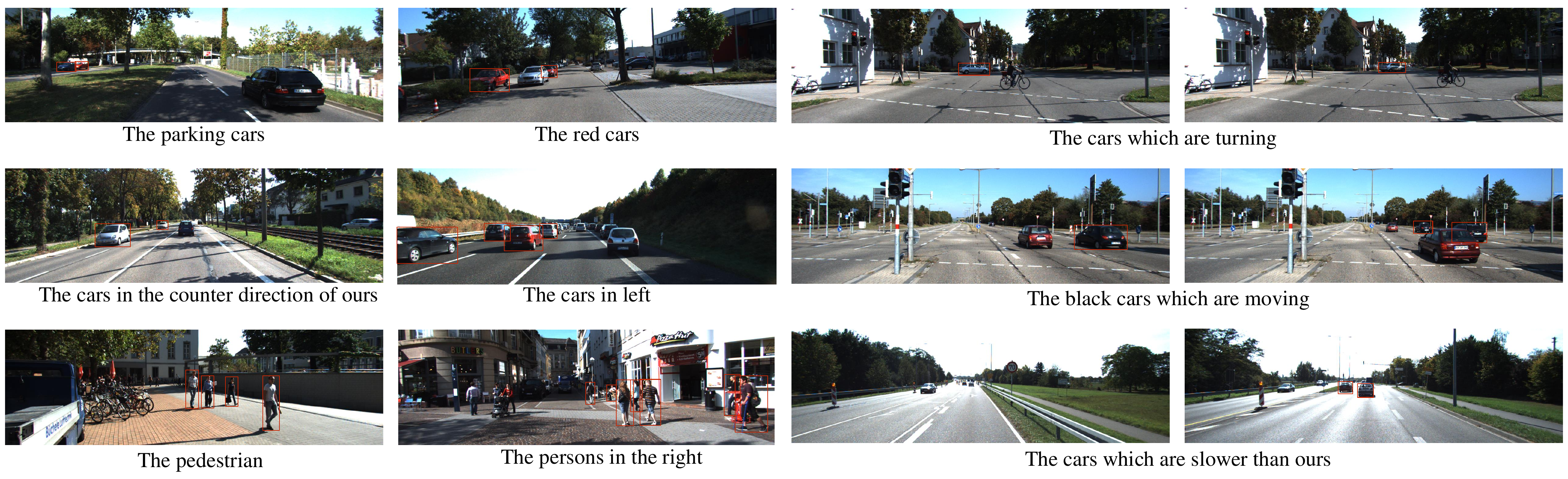}
	}
	\caption{\textbf{Representative examples of Refer-KITTI}.  It provides high-diversity scenes and high-quality annotations referred to by expressions.}
	\label{fig:dataset_examples_rk}
\end{figure*} 
 \begin{figure*}[t]
	\centering
	\resizebox{\textwidth}{!}
	{
		\includegraphics[width = 16cm]{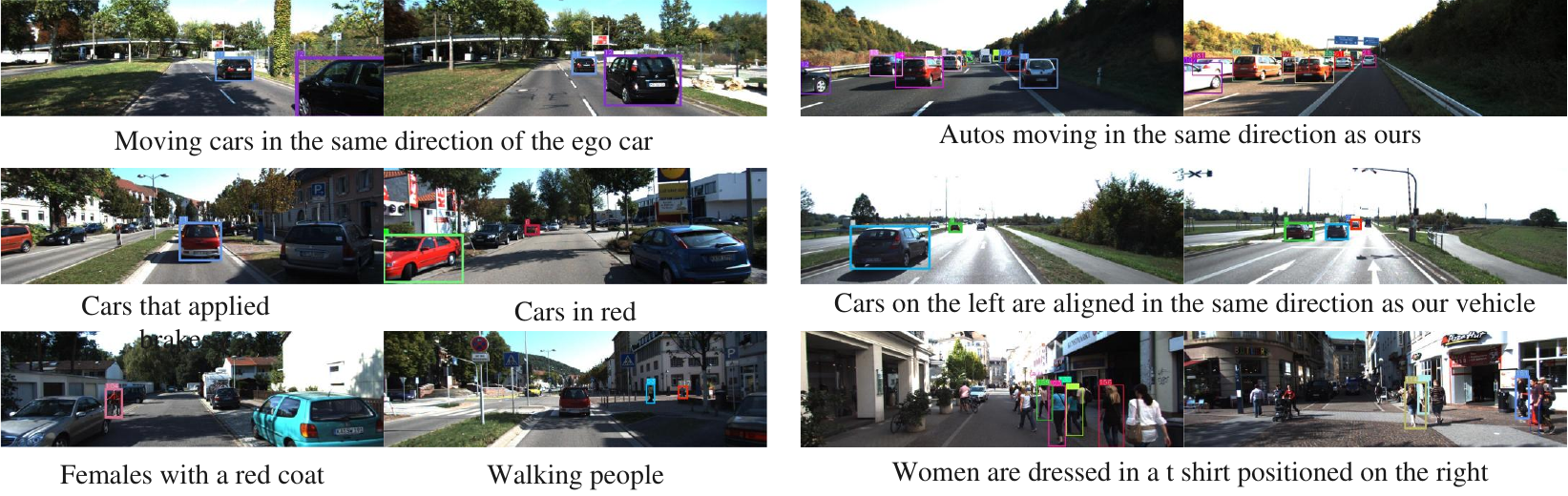}
	}
	\caption{\textbf{Representative examples of Refer-KITTI-V2.} It possesses the advantages of complex motion states, linguistic flexibility, and multi-perspective descriptions, showing strong semantic diversity.}
	\label{fig:dataset_examples_rk2}
\end{figure*} 
\begin{figure*}[t]
  \centering
  \includegraphics[width=1\linewidth]{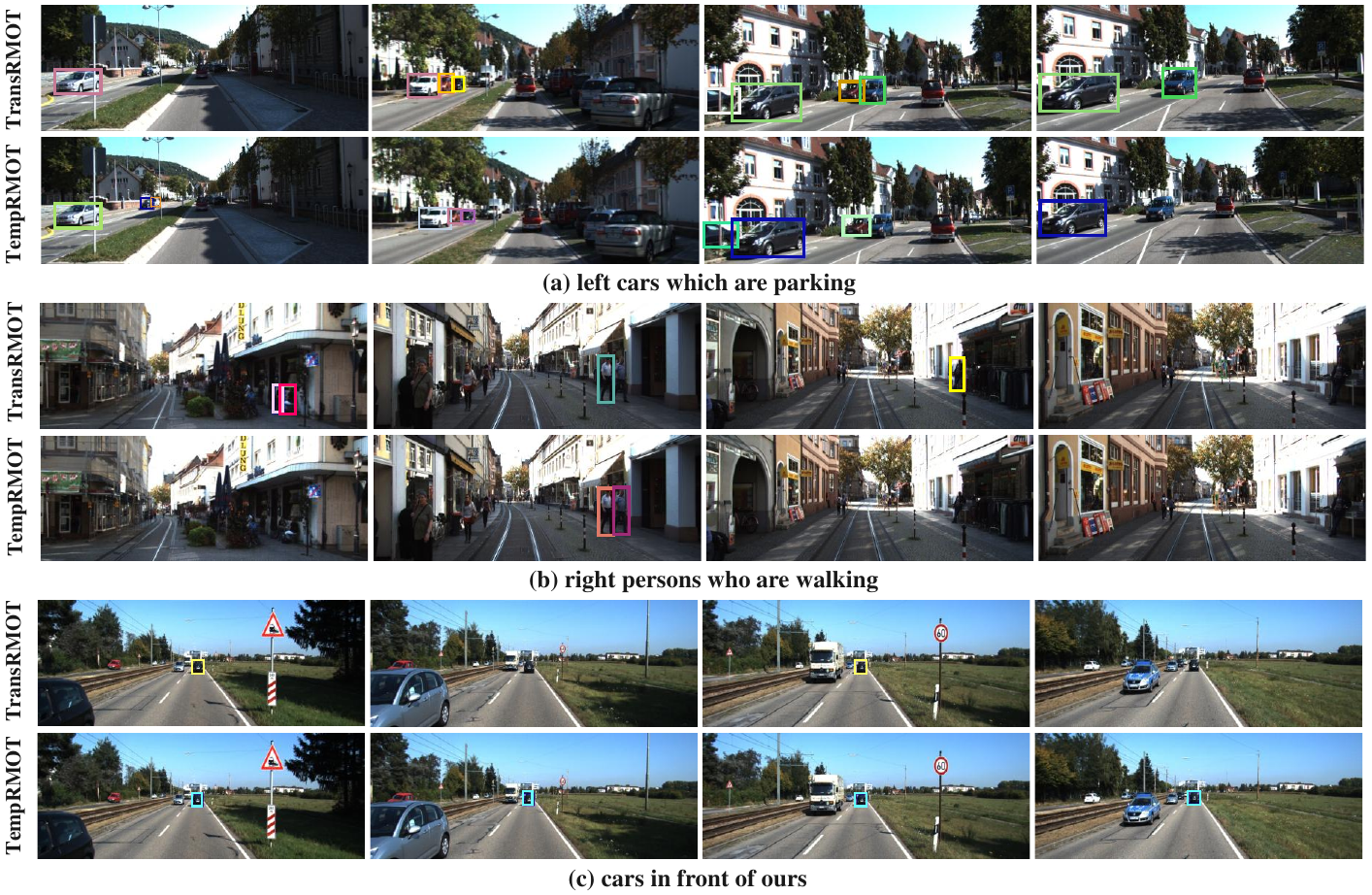}
  \caption{Visualization of predictions from TransRMOT and our TempRMOT. TempRMOT not only accurately understands instructions related to motion without wrong identification, but also successfully detects and tracks objects.}
  \label{fig:qualitative}

\end{figure*}
\begin{table}[]
\centering
\small
\caption{
Comparison with state-of-the-art methods on Refer-BDD and Refer-Dance. The best results are in bold. 
}
\resizebox{1\linewidth}{!}{
\begin{tabular}{p{26mm}|ccccc}
\toprule
\rowcolor[gray]{.9}
Method & HOTA & DetA & AssA & DetRe &	DetPr \\
\midrule
\multicolumn{ 6 }{l}{\textbf{\textit{Refer-BDD dataset}}} \\
MOTRv2~\cite{zhang2023motrv2}&35.35& 22.88 &55.24&\textbf{46.35 }&30.33\\
CO-MOT~\cite{yan2023bridging}&36.72& 24.88& 54.73 &30.01&\textbf{56.95}\\
TransRMOT~\cite{wu2023referring}&34.79& 26.22& 47.56& 38.46 &41.54\\
EchoTrack~\cite{lin2024echotrack} &38.00&28.57&51.24&36.60&54.25 \\
\rowcolor{pink!40}
\rowcolor{pink!40}
\textbf{TempRMOT(Ours)}&\textbf{41.23}&\textbf{30.10}& \textbf{57.91}& 38.09& 56.55 \\

\midrule
\multicolumn{ 6 }{l}{\textbf{\textit{Refer-Dance dataset}}} \\

TransRMOT~\cite{wu2023referring}&9.58&4.37&20.99&6.56&11.55 \\
EchoTrack~\cite{lin2024echotrack}&10.84& 4.73& 24.87& 5.11& 17.94 \\

\rowcolor{pink!40}
\textbf{TempRMOT(Ours)}&\textbf{14.66}& \textbf{8.19}& \textbf{26.32}& \textbf{11.42}& \textbf{22.33} \\
\bottomrule
\end{tabular} 
}
\label{tab:refer-bdd}

\end{table}

\noindent 
\noindent \textbf{Refer-Animal.} Table~\ref{tab:sota_refer_kitti_v2} also reports results on Refer-Animal. TempRMOT also shows significant advantages in this new animal scenario, improving over our conference version~\cite{wu2023referring} across most metrics. This improvement further demonstrates the effectiveness and generalization of TempRMOT.

\noindent 
\noindent \textbf{Refer-BDD.} 
We compare TempRMOT with state-of-the-art methods on the Refer-BDD~\cite{lin2024echotrack} test set, and the results are shown in Table~\ref{tab:refer-bdd}. Without bells and whistles, TempRMOT achieves a HOTA score of 41.23\%, significantly surpassing EchoTrack by 3.23\% HOTA. It also attains the highest detection and association metrics, with DetA of 30.10\% and AssA of 57.91\%. These results demonstrate that TempRMOT effectively leverages visual and linguistic cues to maintain accurate identity tracking, even in complex traffic scenarios with multiple targets and frequent occlusions.

\noindent 
\noindent \textbf{Refer-Dance.} The evaluation on the Refer-Dance test set further validates the effectiveness of TempRMOT for human-centric referring multi-object tracking. As shown in Table~\ref{tab:refer-bdd}, TempRMOT consistently outperforms existing end-to-end methods across all evaluation metrics. Overall, these results demonstrate the strong generalizability of TempRMOT across multiple referring multi-object tracking datasets, highlighting its robustness in both traffic and human-centric scenarios.

\noindent \textbf{KITTI.} 
We further evaluate both TempRMOT and TransRMOT on KITTI, as shown in Table~\ref{tab:sota_refer_kitti_v2}.
Our approach achieves better performance on the HOTA metric, with an improvement of 5.53\% on AssA (72.04\% \textit{vs} 66.51\%), highlighting the effectiveness of our temporal enhanced module.
\subsection{Ablation Study}
\label{sec:ablation}
\noindent \textbf{Module Effectiveness.}
We perform an ablation study on the Refer-KITTI-V2 dataset to assess the contribution of each module in TempRMOT. As presented in Table~\ref{table:module_effective}, the Temporal Enhancement module proves to be the most impactful component. When added to the baseline (a reproduced version of TransRMOT~\cite{wu2023referring}), it yields a substantial HOTA improvement of 3.46\%, underscoring the benefits of incorporating temporal context into the query representation. Furthermore, combining both the Temporal Enhancement module and the Track Refinement module leads to the best overall performance, demonstrating the complementary nature of these components in enhancing tracking precision and temporal consistency.

\noindent \textbf{Effect of Temporal Memory Length During Training.}
TempRMOT employs a temporal memory mechanism to capture and store historical information from the first $N$ moments, enabling the model to reason over extended time spans. To examine the impact of memory length during the training phase, we conduct experiments with varying memory sizes while keeping the length consistent between training and inference. As shown in Table~\ref{table:length_a}, increasing the temporal memory length leads to steady improvements across all evaluation metrics. This suggests that longer historical context allows the model to better capture motion patterns and resolve object ambiguities, which are especially crucial in referring multi-object tracking.
\noindent \textbf{Effect of Temporal Memory Length During Inference.}
We further explore whether increasing the amount of historical information during inference can enhance performance, even when the model is trained with a shorter memory window. Specifically, we fix the memory length to 5 during training and gradually increase it during inference. The results, presented in Table~\ref{table:length_b}, reveal a consistent performance boost as more temporal memory is aggregated at inference time. This highlights the generalization capability of TempRMOT’s temporal module, which can effectively exploit longer temporal cues even beyond those seen during training. The findings underline the importance of temporal modeling for achieving robust and accurate grounding and tracking over extended sequences.
\begin{table}[]
\centering
\small
\caption{
Ablation study on temporal enhancement module. ``Temp.", ``Object." and ``PE" denote the temporal decoder, object decoder, and positional embeddings, respectively.
}
\resizebox{1\linewidth}{!}{
\begin{tabular}{ccc|cccc}
\rowcolor[gray]{.9}
\toprule
Temp. & Object. &PE & HOTA & DetA & AssA& DetRe \\
\midrule
\makecell[c]{\xmarkg} & \makecell[c]{\xmarkg}& \makecell[c]{\xmarkg} &31.00 &19.40 &49.68& \textbf{36.41} \\
\makecell[c]{\cmarkg} & \makecell[c]{\xmarkg}& \makecell[c]{\xmarkg} &33.52 &22.12& 52.86& 32.61 \\
\makecell[c]{\xmarkg} & \makecell[c]{\cmarkg}& \makecell[c]{\xmarkg} & 32.90 &20.64& 52.78& 28.87  \\
\cmarkg & \cmarkg & \makecell[c]{\xmarkg} &34.43&22.28&53.32&32.72\\
\rowcolor{pink!40}
\cmarkg & \cmarkg & \cmarkg &\textbf{35.04}& \textbf{22.97}& \textbf{53.58}& 34.23 \\
\bottomrule
\end{tabular} 
}
\vspace{-10pt}
\label{tab:abation_tempo}
\end{table}
\noindent 
\textbf{Effect of Temporal Enhancement Components.} We also conduct an ablation study to evaluate the contributions of the temporal decoder, object decoder, and positional embeddings in the temporal enhancement module. The results in Table~\ref{tab:abation_tempo} confirm that both modules contribute positively. For example, incorporating only the Temporal Decoder improves HOTA by 2.52\% over the baseline, while the Object Decoder alone also yields an improvement. Their combination leads to further gains, demonstrating their complementary roles.

\noindent 
\textbf{Effect on Implicit Instructions.} Refer-KITTI-V2 introduces implicit instructions involving logical negations or referring to the reference object itself, requiring deeper reasoning (\eg, ``the ego car is positioned after the black cars''). We evaluate models separately on these implicit cases. Results in Table~\ref{tab:implict_expression} show TempRMOT consistently outperforms all competitors, demonstrating its robustness in handling implicit semantics.

\noindent \textbf{Effect of LLM-Generated Expressions.} We evaluated the effect of augmenting human expressions with LLM-generated ones. As Table~\ref{tab:llm_gene} shows, training and evaluating with human and LLM expressions consistently improve performance across all metrics compared to human-only expressions, indicating that LLM-generated expressions provide useful complementary information for detection and association tasks.
\begin{table}[]
\centering
\small
\caption{Comparison on implicit expressions}
\resizebox{\linewidth}{!}{
\begin{tabular}{c|ccccc}
\rowcolor[gray]{.9}
\toprule
Method &HOTA &DetA &AssA&DetRe&DetPr \\
\midrule
iKUN~\cite{du2023ikun}&8.90& 1.61& 49.25& 1.69&24.90  \\

TransRMOT~\cite{wu2023referring}&20.41& 9.47& 44.12& 10.43&\textbf{49.58} \\
\rowcolor{pink!40}
TempRMOT(Ours)&\textbf{22.47}&\textbf{10.13}&\textbf{50.60}&\textbf{12.12}&37.32 \\

\bottomrule
\end{tabular}
}
\label{tab:implict_expression}
\end{table}
\begin{table}[]
\centering
\small
\caption{Effect of LLM-Generated Expressions.}
\resizebox{\linewidth}{!}{
\begin{tabular}{c|ccccc}
\rowcolor[gray]{.9}
\toprule
Expression Type &HOTA &DetA &AssA&DetRe&	DetPr \\
\midrule
Human Only&27.69&15.83& 48.54& 20.88& 38.90 \\
\rowcolor{pink!40}
Human + LLM& \textbf{35.04}& \textbf{22.97}& \textbf{53.58}& \textbf{34.23}& \textbf{40.41} \\

\bottomrule
\end{tabular}
}
\vspace{-2mm}
\label{tab:llm_gene}
\end{table}

\begin{table}[]
\centering
\small
\caption{
Ablation study on temporal enhancement module
}
\resizebox{1\linewidth}{!}{
\begin{tabular}{c|ccccc}
\rowcolor[gray]{.9}
\toprule
Memory Design & HOTA & DetA & AssA& DetRe &	DetPr \\
\midrule
Baseline & 31.00 & 19.40 & 49.68&\textbf{36.41}&28.97 \\
Memory Pooling&31.31& 20.83& 47.21&34.57& 33.89 \\
Convolution &31.52& 22.33& 44.65& 32.09& 41.58 \\
GRU~\cite{cho2014learning}&31.88& 22.57& 45.18& 32.37& \textbf{41.95} \\
\rowcolor{pink!40}
Ours&\textbf{35.04} & \textbf{22.97} & \textbf{53.58} & 34.23&  40.41 \\
\bottomrule
\end{tabular}
}
\label{tab:memory_design}
\end{table}
\noindent \textbf{Effect of Memory Designs.} 
To evaluate the effectiveness of our temporal strategy, we compare the temporal enhancement module with three alternative memory architectures. One uses convolution to aggregate historical query embeddings per object, while the other employs a gated recurrent unit (GRU~\cite{cho2014learning}) to sequentially update queries. As shown in Table~\ref{tab:memory_design}, all temporal strategies provide improvements over the baseline, indicating that leveraging temporal information is generally beneficial. However, our strategy consistently outperforms both alternatives, confirming its efficacy.
\begin{table}[]
\centering
\small
\caption{Ablation study on cross data generalization.}
\resizebox{\linewidth}{!}{
\begin{tabular}{c|c|cccc}
\rowcolor[gray]{.9}
\toprule
Train&Val &HOTA &DetA &AssA \\
\midrule
Refer-Dance~\cite{du2023ikun}&GroOT~\cite{nguyen2024type}&2.46& 0.49& 13.57 \\
Refer-KITTI~\cite{wu2023referring}&GroOT~\cite{nguyen2024type}&6.75& 1.25& 38.69 \\
\rowcolor{pink!40}
Refer-KITTI-V2&GroOT~\cite{nguyen2024type}&
\textbf{12.01}& \textbf{2.91}& \textbf{49.80} \\

\bottomrule
\end{tabular}
}
\vspace{-2mm}
\label{tab:cross_data}
\end{table}
\begin{table}[]
\centering
\small
\caption{Computational cost of the temporal module}
\resizebox{1\linewidth}{!}{
\begin{tabular}{c|cccc}
\rowcolor[gray]{.9}
\toprule
Method & Params & FLOPs & Inference&HOTA \\
\midrule
TempRMOT&49.6M&294.0G&144.18ms&35.04 \\
wo Temp.&42.8M&288.9G&136.77ms&31.00 \\

\bottomrule
\end{tabular}
}
\vspace{-2mm}
\label{tab:temp_cost}
\end{table}
\begin{table}[]
\centering
\small
\caption{Study on Free-Form Human Expressions.}
\resizebox{\linewidth}{!}{
\begin{tabular}{c|ccccc}
\rowcolor[gray]{.9}
\toprule
Expression Type &HOTA &DetA &AssA&DetRe&	DetPr \\
\midrule
Human Only&30.09& 19.09& 47.59& 32.29& 31.41 \\
\rowcolor{pink!40}
Human + LLM& \textbf{37.90}& \textbf{28.34}& \textbf{50.82}& \textbf{39.36}& \textbf{49.35} \\

\bottomrule
\end{tabular}
}
\vspace{-2mm}
\label{tab:human_expression}
\end{table}

\noindent \textbf{Cross Data Generalization.} We evaluate cross-dataset generalization by training on Refer-Dance~\cite{du2023ikun}, Refer-KITTI~\cite{wu2023referring}, and Refer-KITTI-V2, and testing on the MOT17~\cite{milan2016mot16} subset of GroOT~\cite{nguyen2024type}. As shown in Table~\ref{tab:cross_data}, performance improves progressively as the training data becomes more diverse, with the model trained on Refer-KITTI-V2 achieving the highest cross-dataset scores. 
These results demonstrate that, despite substantial domain shifts, training on our datasets generalizes more effectively to unseen scenarios, underscoring the value of diverse annotations in enhancing model generalization.

\noindent \textbf{Computational Cost of the Temporal Module.} We evaluate the computational cost of the temporal enhancement module on the Refer-KITTI-V2 test set, comparing model parameters, forward FLOPs, inference speed, and HOTA with and without the module. As reported in Table~\ref{tab:temp_cost}, the module introduces a small increase in parameters and FLOPs while yielding a clear improvement in HOTA, demonstrating its effectiveness in leveraging temporal context for multi-object tracking.

\noindent \textbf{Generalization to Independently Written Expressions.} To examine potential distributional bias from LLM-generated expressions, two annotators independently write 80 expressions for four test videos without rule-based matching or LLM assistance, and all expressions are verified for correctness. This test set therefore better reflects real-world language use. We then compare models trained on the Refer-KITTI-v2 training set with Human-only and Human+LLM expressions. As shown in Table~\ref{tab:human_expression}, Human+LLM achieves a HOTA score of 37.90, compared with 30.09 for Human-only. This improvement suggests that LLM augmentation increases the diversity of training expressions and improves generalization without introducing an evident bias toward LLM-specific patterns.

\subsection{Qualitative Results}
\label{sec:qulitative}
\begin{figure}[t]
	\centering
	\includegraphics[width=1\linewidth]{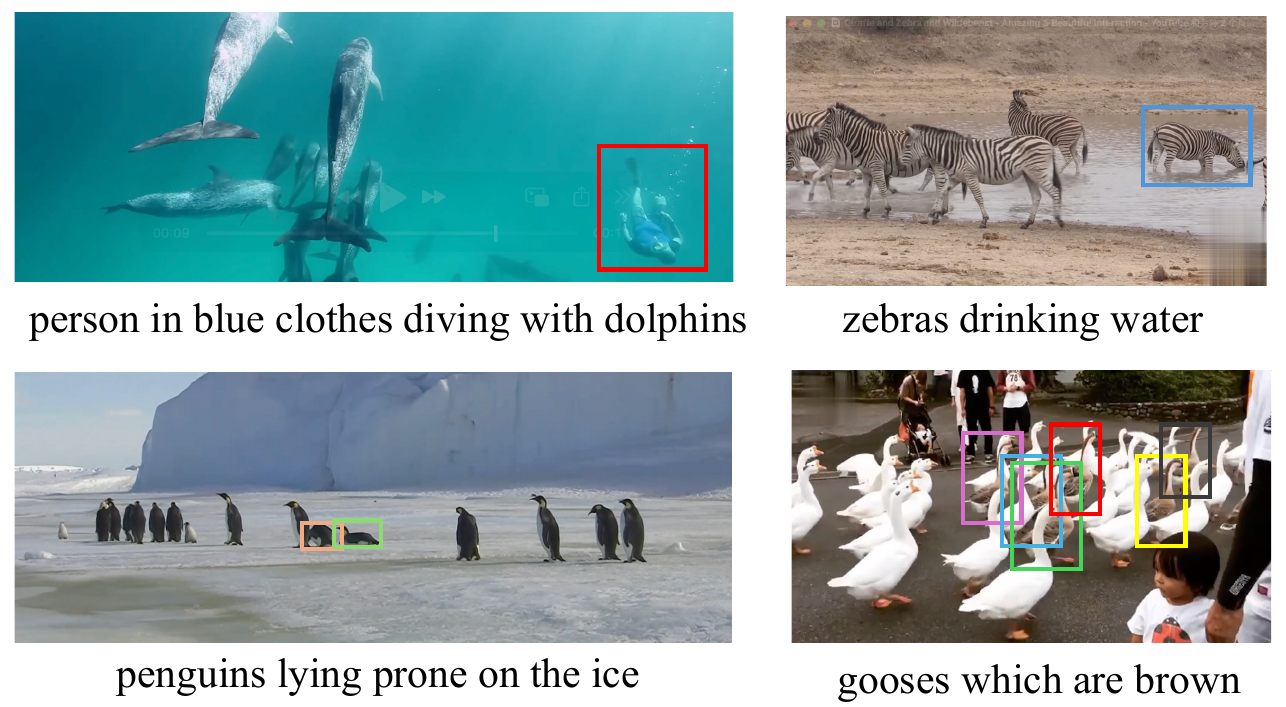}
	\caption{ \textbf{Representative examples of Refer-Animal.} 
	}
	\label{fig:refer-animal}
\vspace{-8pt}
\end{figure}
We present several representative qualitative results in Figure~\ref{fig:qualitative}. As illustrated, TempRMOT effectively interprets complex instructions involving motion and demonstrates accurate object tracking. This indicates that our model can distinguish between objects of the same category based on their motion states.
For example, in Figure~\ref{fig:qualitative}(a), TransRMOT misclassifies moving pedestrians as parked cars, whereas TempRMOT correctly identifies and tracks the stationary vehicles. A similar pattern is observed for the instruction ``right persons who are walking'', where TempRMOT exhibits superior understanding of motion context.
Furthermore, by leveraging historical information, TempRMOT performs more robust tracking and alleviates tracking failures. For example, given the instruction ``cars in front of ours'', TransRMOT fails to maintain consistent object identities and loses track of relevant targets, whereas TempRMOT preserves accurate tracking across frames.

\noindent \textbf{Representative examples of Refer-KITTI.}
Representative examples of Refer-KITTI can be found in Figure~\ref{fig:dataset_examples_rk}, which shows a wide variety of complex driving scenes and diverse object interactions. It also highlights the high-quality annotations provided for each frame, which include detailed referring expressions aligned with corresponding visual targets.

\noindent \textbf{Representative examples of Refer-KITTI-V2.}
We also present representative examples from Refer-KITTI-V2 in Figure~\ref{fig:dataset_examples_rk2}. As illustrated, the dataset contains diverse referring expressions that capture rich and fine-grained semantics. For instance, the expression ``cars in red'' emphasizes both the color and type of vehicles. Another example, ``autos moving in the same direction as ours'', specifies both the action and moving direction of the referents. Similarly, the phrase ``women are dressed in a t-shirt positioned on the right'' conveys information regarding position, gender, clothing, and plurality. Moreover, concise expressions such as ``females with a red coat'' provide detailed descriptions based on appearance. These examples collectively demonstrate the semantic richness and descriptive specificity of Refer-KITTI-V2, which facilitates more precise understanding and interpretation of objects and actions in complex autonomous driving scenarios. Representative examples of Refer-Animal can also be found in Figure~\ref{fig:refer-animal}. These examples demonstrate visual diversity beyond autonomous driving scenarios.

\subsection{Discussion.}
\label{sec:discussion}
\begin{figure}[t]
	\centering
	\includegraphics[width=1\linewidth]{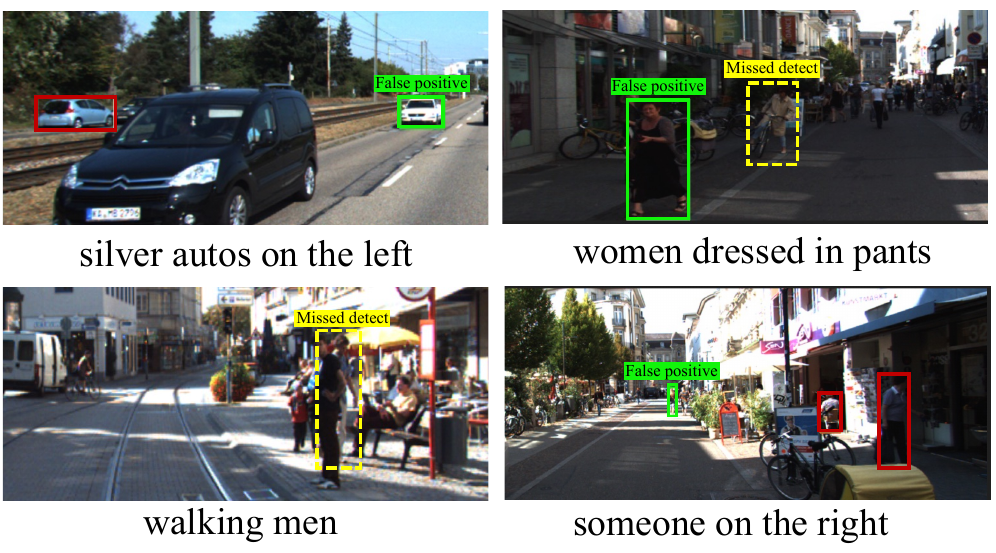}
    \caption{ \textbf{Failure cases of TempRMOT.} 
	}
	\label{fig:failed_cases}
\vspace{-8pt}
\end{figure}

\noindent 
Despite its strong overall performance, TempRMOT still faces three challenges illustrated in Figure~\ref{fig:failed_cases}.
(1) Limited fine-grained attribute discrimination can cause confusion between similar colors or clothing.  (2) In the ``walking men'' example, the target first appears under severe occlusion. Incomplete visual evidence and the lack of historical queries can therefore lead to missed detections. (3) Small objects provide limited visual details, reducing the reliability of language matching. Future work will focus on improving fine-grained discrimination and robustness to occlusion and small-scale objects.

\section{Conclusion}
\label{sec:conclusion}
In this paper, we proposed a novel referring understanding task, called Referring Multi-Object Tracking (RMOT). 
It addressed the single-object limitation of referring understanding tasks and provided a more flexible multi-object setting. 
To facilitate research on RMOT, we developed three benchmarks, named Refer-KITTI, Refer-KITTI-V2, and Refer-Animal.
Refer-KITTI is characterized by high flexibility in referent object selection and pronounced temporal dynamics, while maintaining low annotation costs.
Refer-KITTI-V2 provided a more diverse and larger-scale dataset, with 9,758 expressions and 617 unique words. 
It addressed the limitations of previous benchmarks by introducing more expressions, richer semantics, and implicit expressions. 
Refer-KITTI and Refer-KITTI-V2 focus on autonomous driving scenarios, while Refer-Animal targets animal scenes.
Additionally, we introduce TempRMOT, a query-based framework for long-term spatial-temporal modeling. By retrieving historical queries and integrating current-frame object interactions, it better captures object dynamics and interprets motion-related expressions.

{\small
\bibliographystyle{unsrt}
\bibliography{rmot}
}
\begin{IEEEbiographynophoto}
    {Yani Zhang} received the BS degree in the School of Computer Science from Wuhan University in 2024. She is currently working toward the ME degree with the School of Computer Science, Wuhan University. Her current research interests include vision-language learning, embodied perception and multi-modal LLMs.
\end{IEEEbiographynophoto}

\begin{IEEEbiographynophoto}
{Dongming Wu} is currently a Postdoctoral Fellow with the Multimedia Lab (MMLab) at the Chinese University of Hong Kong. 
He received the Ph.D. degree in Computer Science from Beijing Institute of Technology in 2025.
His current research interests include vision-language learning, and multi-modal LLMs.
He has published more than 10 papers in refereed conference proceedings and journals such as IEEE Transactions on Pattern Analysis and Machine Intelligence (TPAMI), ICML, ICLR, CVPR, ICCV, ECCV, etc.
\end{IEEEbiographynophoto}

\begin{IEEEbiographynophoto}
{Wencheng Han} is currently a Full Professor with the School of Control Science and Engineering, Shandong University, Jinan, China. He received the Ph.D. degree in computer science from the University of Macau, Macao, China, in 2025, where he subsequently worked as a Postdoctoral Fellow with the State Key Laboratory of Internet of Things for Smart City (SKL-IOTSC) from 2025 to 2026. His research interests include computer vision, autonomous driving, robot control, and artificial intelligence-generated content (AIGC). He has authored or co-authored more than 30 papers in leading journals and conferences, including IEEE Transactions on Pattern Analysis and Machine Intelligence (TPAMI), CVPR, ICCV, and ECCV.
\end{IEEEbiographynophoto}

\begin{IEEEbiographynophoto}
{Xingping Dong} received the BS degree from Xiamen University, in 2012, and the Ph.D. degree from Beijing Institute of Technology, in 2019. He is currently a professor with the School of Computer Science, Wuhan University. Before that, he was a research scientist with the Inception Institute of Artificial Intelligence (IIAI), UAE. From 2016 to 2018, he was a joint PhD candidate with the College of Engineering and Computer Science, Australian National University. His current research interests include computer vision, autonomous driving, and deep learning.
\end{IEEEbiographynophoto}

\begin{IEEEbiographynophoto}
{Shengcai Liao} is an Associate Professor in the College of Computing and Artificial Intelligence (CCAI) in UAEU. He is an IEEE Fellow and IAPR Fellow. He was a Principal Scientist in IIAI, G42 in UAE during 2018-2024, and an Associate Professor in Institute of Automation, Chinese Academy of Sciences (CASIA) from 2014 to 2018. He received B.S. degree in mathematics from Sun Yat-sen University in 2005 and Ph.D. degree from CASIA in 2010. He was a Postdoc in Michigan State University during 2010-2012. His research interests include object detection, recognition, image generation, especially face and person analysis and synthesis. He was awarded Best Paper/Best Student Paper in ICB 2006/2007/2015, and CCBR 2016, Best/Outstanding Reviewer in IJCB 2014 and CVPR 2019/2021. He served as a Program Chair for IJCB 2022, Area Chair for ICPR 2016, ICB 2016/2018, CVPR 2022/2023, ICCV 2023, and NeurIPS 2023.
\end{IEEEbiographynophoto}

\begin{IEEEbiographynophoto}
{Bo Du} (M’10–SM’15)  is currently a professor with the School of Computer Science and Institute of Artificial Intelligence, Wuhan University. He is also the director of National Engineering Research Center for Multimedia Software, Wuhan University, Wuhan, China. He has more than 80 research papers published in the IEEE Transactions on Image Processing (TIP), IEEE Transactions on Cybernetics (TCYB), IEEE Transactions on Pattern Analysis and Machine Intelligence(TPAMI), IEEE Transactions on Geoscience and Remote Sensing (TGRS), etc. Twenty-four of them are ESI hot papers or highly cited papers. His major research interests include machine learning, computer vision, and image processing. He won IEEE Geoscience and Remote Sensing Society 2020 Transactions Prize Paper Award. He won the IJCAI (International Joint Conferences on Artificial Intelligence) Distinguished Paper Prize, IEEE Data Fusion Contest Champion, and IEEE Workshop on Hyperspectral Image and Signal Processing Best Paper Award in 2018. He regularly serves as a senior PC member of IJCAI and AAAI. He served as area chair for ICPR.
\end{IEEEbiographynophoto}
\end{document}